\documentclass[journal]{IEEEtran}

\usepackage{scalerel}
\usepackage{tikz}
\usetikzlibrary{svg.path}
\usepackage{xcolor}
\usepackage{amsmath}
\usepackage{amsfonts}
\usepackage{amssymb}
\usepackage{mathtools}
\usepackage[numbers]{natbib}
\usepackage[ruled,vlined,noend,linesnumbered]{algorithm2e}
\usepackage{caption}
\usepackage{subcaption}
\usepackage{multirow}
\usepackage{booktabs}
\usepackage{array}
\usepackage{blkarray}
\usepackage[utf8]{inputenc}
\usepackage{fancyhdr}
\usepackage{pifont}
\newcommand{\cmark}{\ding{51}}%
\newcommand{\xmark}{\ding{55}}

\usepackage{contour}
\usepackage{ulem}
\contourlength{0.8pt}

\newcommand{\myuline}[1]{%
  \uline{\phantom{#1}}%
  \llap{\contour{white}{#1}}%
}

\definecolor{orcidlogocol}{HTML}{A6CE39}
\tikzset{
    orcidlogo/.pic={
        \fill[orcidlogocol] svg{M256,128c0,70.7-57.3,128-128,128C57.3,256,0,198.7,0,128C0,57.3,57.3,0,128,0C198.7,0,256,57.3,256,128z};
        \fill[white] svg{M86.3,186.2H70.9V79.1h15.4v48.4V186.2z}
        svg{M108.9,79.1h41.6c39.6,0,57,28.3,57,53.6c0,27.5-21.5,53.6-56.8,53.6h-41.8V79.1z M124.3,172.4h24.5c34.9,0,42.9-26.5,42.9-39.7c0-21.5-13.7-39.7-43.7-39.7h-23.7V172.4z}
        svg{M88.7,56.8c0,5.5-4.5,10.1-10.1,10.1c-5.6,0-10.1-4.6-10.1-10.1c0-5.6,4.5-10.1,10.1-10.1C84.2,46.7,88.7,51.3,88.7,56.8z};
    }
}

\newcommand\orcidicon[1]{\href{https://orcid.org/#1}{\mbox{\scalerel*{
                \begin{tikzpicture}[yscale=-1,transform shape]
                \pic{orcidlogo};
                \end{tikzpicture}
            }{|}}}}

\usepackage{hyperref}

\def\@fnsymbol#1{\ensuremath{\ifcase#1\or *\or \dagger\or \ddagger\or
   \mathsection\or \mathparagraph\or \|\or **\or \dagger\dagger
   \or \ddagger\ddagger \else\@ctrerr\fi}}
\newcommand{\ssymbol}[1]{^{\@fnsymbol{#1}}}

\begin{document}
\title{TransCODE: Co-design of Transformers and Accelerators for Efficient Training and Inference}

\author{Shikhar~Tuli$^{\textsuperscript{\orcidicon{0000-0002-9230-5877}}}$,~\IEEEmembership{Student Member,~IEEE,} and~Niraj~K.~Jha,~\IEEEmembership{Fellow,~IEEE}
\thanks{This work was supported by NSF Grant No. CNS-2216746. S. Tuli and N. K. Jha are with the Department of Electrical and Computer Engineering,
Princeton University, Princeton, NJ, 08544, USA (e-mail: \{stuli, jha\}@princeton.edu).}
\thanks{Manuscript received ---; revised ---.}}

\markboth{}{Tuli \MakeLowercase{\textit{et al.}}: TransCODE: Co-design of Transformers and Accelerators for Efficient Training and Inference}


\maketitle

\begin{abstract}

Automated co-design of machine learning models and evaluation hardware is critical for efficiently deploying 
such models at scale. Despite the state-of-the-art performance of transformer models, they are not yet ready
for execution on resource-constrained hardware platforms. High memory requirements and low parallelizability of 
the transformer architecture exacerbate this problem. Recently-proposed accelerators attempt to optimize 
the throughput and energy consumption of transformer models. However, such works are either limited to a 
one-sided search of the model architecture or a restricted set of off-the-shelf devices. Furthermore, previous 
works only accelerate model inference and not training, which incurs substantially higher memory and compute 
resources, making the problem even more challenging. To address these limitations, this work proposes a 
dynamic training framework, called DynaProp, that speeds up the training process and reduces memory consumption. 
DynaProp is a low-overhead pruning method that prunes activations and gradients at runtime. To effectively execute 
this method on hardware for a diverse set of transformer architectures, we propose ELECTOR, a framework that
simulates transformer inference and training on a design space of accelerators. We use this simulator in conjunction 
with the proposed co-design technique, called TransCODE, to obtain the best-performing models with high accuracy on 
the given task and minimize latency, energy consumption, and chip area. The obtained transformer-accelerator pair 
achieves 0.3\% higher accuracy than the state-of-the-art pair while incurring 5.2$\times$ lower latency and 
3.0$\times$ lower energy consumption.

\end{abstract}


%
\IEEEpeerreviewmaketitle

\section{Introduction}

\IEEEPARstart{A}{rtificial} intelligence (AI) is undergoing a paradigm shift with the rise of large language models, 
e.g., BERT~\cite{bert}, GPT-3~\cite{gpt_3}, DALL-E~\cite{dall-e}). These models, backed by the transformer 
architecture~\cite{vaswani}, target many applications, including language~\cite{bert}, vision~\cite{vit_2021}, robotic 
manipulation~\cite{txf_robotics}, reasoning~\cite{txf_proof}, human interaction~\cite{txf_hi}, and 
forecasting~\cite{txf_forecasting}. Training on comprehensive datasets (generally using self-supervision at scale) and 
finetuning on downstream tasks have enabled their widespread application. However, training and inference with such large 
models either involve high power consumption on graphical processing units (GPUs) or high energy and latency on 
off-the-shelf edge-AI devices. For instance, the lowest possible latency for transformer inference on a Raspberry 
Pi~\cite{rpi} is 2.1 seconds~\cite{edgetran}, which is too slow for real-time natural language processing (NLP) tasks. 
This makes efficient training (and even inference) of such models an unsolved problem. The increasing size of 
state-of-the-art language models~\cite{gpt_3, turing_nlg} results in a higher memory footprint and computational 
complexity, exacerbating this problem.

Previous works propose many specialized hardware accelerators to address the abovementioned challenges. For instance, 
A$^3$~\cite{a3} is one of the first accelerators to enable efficient transformer inference by leveraging algorithmic 
approximation and hardware specialization. SpAtten~\cite{spatten} proposes a cascade token pruning mechanism to prune 
the weights of a transformer at runtime. Energon~\cite{energon} approximates this pruning mechanism to speed up inference. 
AccelTran~\cite{acceltran}, a state-of-the-art transformer accelerator, executes dynamic inference by skipping 
\emph{all} ineffectual multiply-and-accumulate (MAC) operations. It also \emph{tiles} the matrices to facilitate higher 
parallelization and hardware utilization while executing the matrix multiplication operations. However, the above 
accelerators do not support transformer training, which demands a higher memory footprint for transformer execution 
on resource-constrained edge-AI devices. 

Tuli et al.~\cite{flexibert} showed that each NLP task has a unique optimal transformer architecture, requiring a 
specialized accelerator for efficient evaluation. However, designing a single accelerator that efficiently executes a 
diverse set of transformer architectures takes significant time and effort. Transformers involve serially-connected 
encoder (and sometimes decoder) layers. An accelerator with too many processing elements (PEs) would have low compute 
resource utilization for a deep but narrow transformer model. A PE is the basic compute module in an accelerator. A 
shallow and wide transformer would incur low latency and enable high parallelization~\cite{edgetran, hat_mit} but would 
require many PEs and high bandwidth memory. However, a manual search of the best accelerator architecture and transformer 
design decisions is computationally too expensive due to the vastness of each design space~\cite{codebench}.

To tackle the abovementioned challenges, previous works implement automated hardware-software co-design. NAAS~\cite{naas} 
and CODEBench~\cite{codebench} simultaneously search for the best design choices for the convolutional neural network (CNN) 
and accelerator architecture. However, a CNN workflow is different from that of a transformer, warranting substantially 
different accelerator design choices. Some recent works target co-design with transformer models. Qi et 
al.~\cite{qi_iccad_21} use a recurrent neural network (RNN) and a reinforcement learning (RL)-based controller to guide 
the search using a pool of five field-programmable gate arrays (FPGAs) and adjust the pruning parameters of an input 
transformer model. Peng et al.~\cite{peng_dac_22} explore the scheduling and sparsity decisions on a single FPGA. 
CODEBench~\cite{codebench}, although a framework for CNN accelerators, shows the advantages of exploring massive CNN and 
accelerator search spaces, resulting in high gains in accuracy, energy consumption, evaluation throughput, and chip area. \textcolor{black}{Hence, we leverage its co-design technique, BOSHCODE, in our proposed framework (details in Section~\ref{sec:hw_co_design}).} 
On the other hand, the abovementioned works on transformer accelerator search only target one (or few) transformer models 
on a limited set of hardware platforms. This restricts the gains from automated co-design, leading to low resource 
utilization and inefficient configurations.

In order to address the above issues, we propose TransCODE, a co-design framework for transformers and 
application-specific integrated circuit (ASIC)-based accelerators. Our main contributions are as follows.

\begin{itemize}
    \item For efficient on-device training, we propose DynaProp, which dynamically prunes weights, activations, and 
gradients to skip ineffectual MAC operations and speed up the transformer training/inference process. DynaProp 
leverages specialized low-overhead hardware modules to induce sparsity into transformer training and inference.
    \item To support vast design spaces involving \emph{flexible} and \emph{heterogeneous} transformer 
architectures~\cite{flexibert}, we propose a fl\underline{e}xib\underline{l}e B\underline{E}RT 
a\underline{c}celera\underline{tor} (ELECTOR) framework. ELECTOR supports diverse transformer architectures within the 
FlexiBERT 2.0 design space~\cite{edgetran}. It efficiently implements model operations through dedicated hardware modules 
and a functional transformer mapper. The design space within the ELECTOR framework involves disparate accelerators that 
can execute the transformers in the FlexiBERT 2.0 design space. ELECTOR also effectively implements the proposed 
DynaProp algorithm to speed up transformer training and inference. It involves 14,850,000 accelerators, a design space much 
more extensive than investigated in any previous work.
    \item We then leverage the proposed ELECTOR and FlexiBERT 2.0 design spaces to implement co-design and obtain a 
transformer-accelerator pair that maximizes the performance objectives within the given user-defined constraints. We 
call this framework, which co-designs the transformer-accelerator pair, TransCODE. It leverages the best-performing 
optimization technique in the considered design spaces.
\end{itemize} 

We organize the rest of the article as follows. Section~\ref{sec:background} presents background on transformer and 
accelerator design choices along with automated hardware-software co-design. Section~\ref{sec:methodology} illustrates 
the TransCODE framework that includes DynaProp, ELECTOR framework and its design space, and the co-design pipeline. 
Section~\ref{sec:exp_setup} describes the experimental setup and targeted baselines. Section~\ref{sec:results} 
presents the results. \textcolor{black}{Section~\ref{sec:discussion} discusses the limitations of the proposed work and future work directions.} Finally, Section~\ref{sec:conclusion} concludes the article.

\section{Background and Related Work}
\label{sec:background}

In this section, we present background material on popular transformer and accelerator architectures and the 
corresponding design decisions. We also describe previously proposed hardware-software co-design methods.

\subsection{Transformer Design Space}

Previous works propose various transformer architectures. BERT is one of the most popular architectures that is widely 
used for language modeling~\cite{bert}. Its variants leverage mechanisms other than vanilla 
self-attention~\cite{shaw2018selfattention} to optimize performance or reduce model size and complexity. They include 
RoBERTa~\cite{roberta} that implements robust pre-training techniques, ConvBERT~\cite{convbert} 
that uses one-dimensional convolutional operations, MobileBERT~\cite{mobilebert} that employs bottleneck structures and 
multiple feed-forward stacks, among many others. Further, architectures like FNet~\cite{fnet} and 
LinFormer~\cite{linformer} use Fourier transform and low-rank approximation, respectively, of the self-attention 
operation to aid efficiency and reduce the number of model parameters. 

In order to search for the best-performing model for a given task, FlexiBERT~\cite{flexibert} unifies and 
implements \emph{heterogeneous} and \emph{flexible} transformer architectures, encapsulating various self-attention 
operation types. Each encoder layer in its design space can have a different attention mechanism (heterogeneity) and a 
different hidden dimension (flexibility). Among many works that implement neural architecture search (NAS) in a 
design space of transformer models~\cite{schubert, dynabert, autotinybert, nas-bert}, FlexiBERT has the largest and 
the most expressive design space. This results in state-of-the-art models that outperform previous architectures in 
accuracy. FlexiBERT 2.0~\cite{edgetran} extends the design space to 1.7 $\times$ 10$^{88}$ transformer 
models, the largest and the most expressive transformer design space to date. We thus use the FlexiBERT 2.0 design space 
to implement co-design in this work. Note that no previously proposed accelerator supports heterogeneous and flexible 
transformer workflows. We discuss traditional transformer accelerators next.

\subsection{Accelerator Design Space}

\textcolor{black}{A transformer model's hardware performance (characterized by latency, energy consumption, and chip area) on 
a given platform depends on multiple factors. These factors include memory size and bandwidth, number of MAC units (that can 
execute matrix multiplication operations in parallel), number of specialized hardware modules (e.g., ones for softmax and 
layer-norm operations), operation scheduling, dataflow, model sparsity, etc. These design decisions lead to many existing 
accelerators proposed in the literature.}

A$^3$~\cite{a3} is one of the first ASICs to support transformer acceleration. It uses several approximation strategies 
to avoid computing attention scores that are close to zero. SpAtten~\cite{spatten} proposes the top-$k$ pruning 
algorithm that ranks input token and attention-head scores using a dedicated hardware module. However, it only considers 
part of the activations formed, not sparsity in all possible matrix multiplication operations. Further, implementing 
the proposed top-$k$ pruning mechanism involves high compute overhead; its time complexity is $\mathcal{O}(N^3)$, leading 
to marginal gains in energy efficiency~\cite{acceltran}. Energon~\cite{energon} approximates this pruning mechanism. 
However, since it is limited to being a co-processor, it requires high off-chip memory access. Finally, 
OPTIMUS~\cite{optimus} targets sparsity in a broader scope, using a set-associative rearranged compressed sparse column 
format to eliminate ineffectual MAC operations, although limited to weight matrices. Here, weights correspond to the 
trainable transformer model parameters and activations are represented by intermediate matrices formed by the 
transformer model operations.

To overcome the drawbacks of the abovementioned accelerators, AccelTran~\cite{acceltran} implements dynamic inference 
with a transformer while pruning \emph{all} weights and activations. In addition, it leverages matrix tiling 
to improve parallelization and resource utilization. However, it only executes transformer inference and not training, 
uses a fixed set of design choices (e.g., a fixed tile size, number of PEs, buffer sizes), and does not support diverse 
models, thus leading to sub-optimal utilization. To tackle this problem, various works propose design spaces of transformer 
accelerators to efficiently obtain the optimal transformer architecture for the given task. However, such design spaces 
are limited to off-the-shelf FPGAs~\cite{qi_iccad_21, peng_dac_22} that only focus on inference. We next describe previous 
works on co-design of the AI model and hardware accelerator.

\subsection{Hardware-software Co-design}
\label{sec:hw_co_design}

Various works target CNN-accelerator co-design~\cite{codebench, naas, bobw}. CODEBench~\cite{codebench} searches over 
massive CNN and accelerator design spaces. However, its accelerators are geared toward CNN workflows and thus inefficient 
for transformer pipelines. As discussed before, Qi et al.~\cite{qi_iccad_21} use an RNN and RL-based controller 
to guide search in a pool of five FPGAs and adjust the pruning parameters of an input transformer model. However, they 
only consider latency and accuracy constraints and do not optimize energy consumption and chip area. Peng et 
al.~\cite{peng_dac_22} explore the scheduling and sparsity decisions on an FPGA and adapt the input sequence length. 
SpAtten~\cite{spatten} implements hardware-aware NAS (HW-NAS), where it finds a sub-net of a trained 
super-net~\cite{hat_mit}. However, its design space only involves 72 transformers that are not flexible. Thus, there is 
a need for an exhaustive design space of transformer accelerators to implement co-design and obtain the best-performing 
transformer-accelerator pair. This pair should not only deliver high accuracy on a given task but also be 
energy-efficient and have a high throughput and low chip area.

In this work, we leverage \underline{B}ayesian \underline{o}ptimization using \underline{s}econd-order gradients and 
a \underline{h}eteroscedastic surrogate model for \underline{co}-\underline{de}sign, i.e., BOSHCODE~\cite{codebench}. 
It is a scalable co-design framework that efficiently searches the hardware and software design spaces at scale. 
CODEBench~\cite{codebench} proposes and uses BOSHCODE to search over significantly large design spaces 
(4.2 $\times$ 10$^{812}$ CNNs and 2.3 $\times$ 10$^8$ accelerators). EdgeTran~\cite{edgetran} leverages BOSHCODE to 
search over the joint space of FlexiBERT 2.0 and a set of off-the-shelf edge-AI devices, including Raspberry 
Pi~\cite{rpi}, Apple M1~\cite{apple_m1} system-on-chip (SoC) with a central processing unit, CPU, and a GPU, Intel 
Neural Compute stick~\cite{ncs} (a neural processing unit), and Nvidia Jetson Nano~\cite{nano} (SoC with both CPU and 
GPU). 

BOSHCODE supports co-design with any two search spaces. It leverages second-order gradient-based 
optimization~\cite{tuli2021cosco} on an actively-trained~\cite{al_survey} surrogate model for performance 
prediction (which is the optimization objective). The surrogate model combines a natural parameter network (NPN), 
a teacher, and a student network. The NPN predicts the mean performance of the transformer-accelerator pair along with 
the aleatoric uncertainty. The teacher and student networks predict the epistemic uncertainty in performance. Epistemic 
uncertainty is the uncertainty in performance due to an unexplored design space. In contrast, aleatoric uncertainty refers 
to the uncertainty due to parameter initializations and variations in model performance due to different training recipes. 
BOSHCODE exploits epistemic and aleatoric uncertainty estimates to obtain the best design decisions for the transformer, 
the accelerator, and the model training recipe that maximizes accuracy. We present more details on how we leverage 
BOSHCODE in our search process in Section~\ref{sec:transcode}.

\section{Methodology}
\label{sec:methodology}

\begin{figure*}
    \centering
    \includegraphics[width=0.9\linewidth]{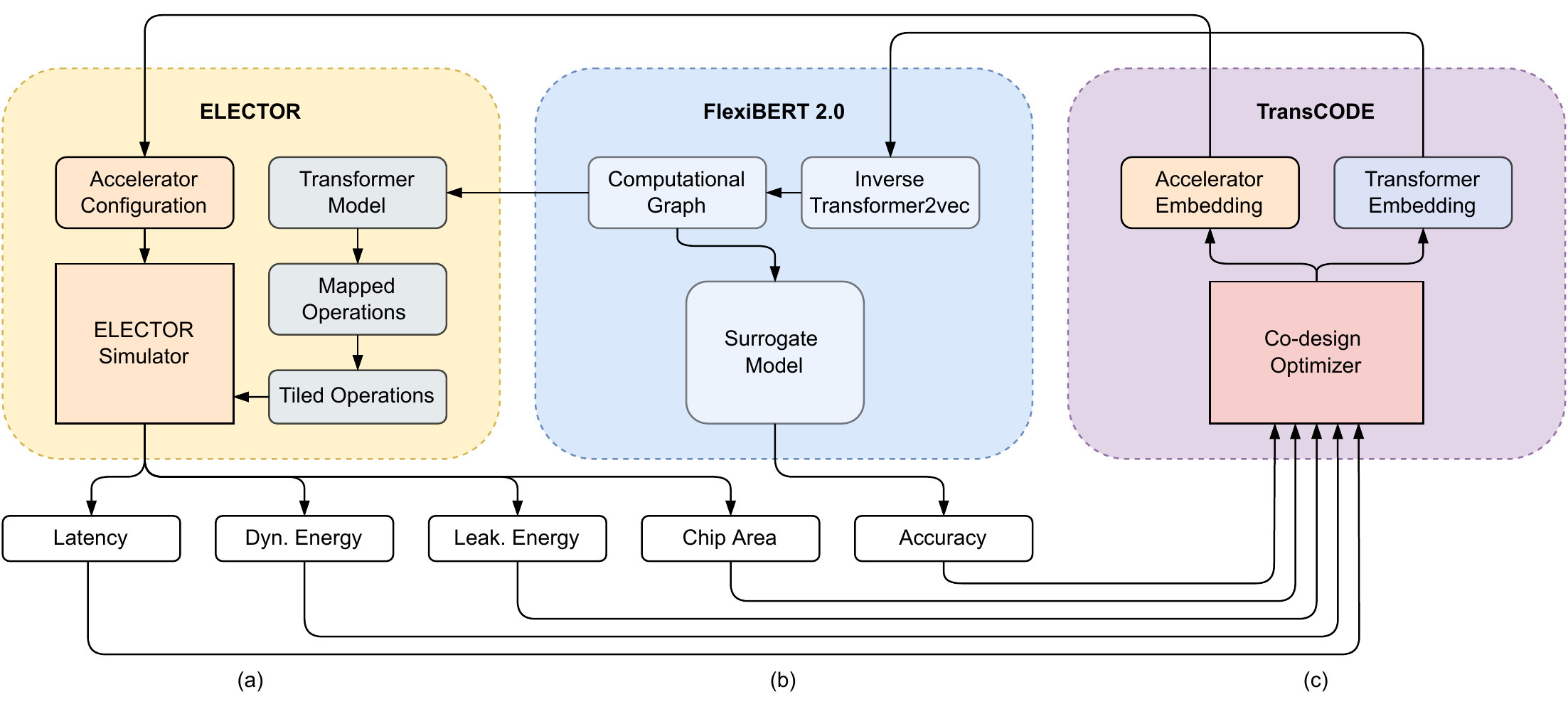}
    \caption{Overview of the TransCODE framework. (a) ELECTOR takes an accelerator embedding and a transformer 
computational graph to simulate its training/inference on the given accelerator. (b) FlexiBERT 2.0 converts the 
input transformer embedding to a computational graph and employs a pre-trained surrogate model to predict model 
accuracy. (c) The TransCODE optimizer takes in the performance values of the previously evaluated 
transformer-accelerator pair to query another pair in the active learning loop.}
    \label{fig:transcode_flowchart}
\end{figure*}

Fig.~\ref{fig:transcode_flowchart} shows an overview of the TransCODE framework. ELECTOR, in 
Fig.~\ref{fig:transcode_flowchart}(a), takes the accelerator embedding and transformer computational graph as input. 
Using the accelerator embedding, it implements a hardware accelerator with the corresponding design decisions. Next, 
it converts the computational graph into a corresponding transformer model with modular operations (supported by the 
FlexiBERT 2.0 design space), which it then maps to specialized hardware modules. It also tiles the matrices for efficient 
resource allocation, operation scheduling, and data reuse. Fig.~\ref{fig:transcode_flowchart}(b) shows how we leverage 
the FlexiBERT 2.0~\cite{flexibert} framework to convert a transformer embedding to its corresponding computational graph 
and employ the surrogate model to predict model accuracy. Finally, Fig.~\ref{fig:transcode_flowchart}(c) illustrates 
TransCODE, which uses previous performance results to train a surrogate model and query the next transformer-accelerator 
pair. Finally, it feeds the output accelerator and transformer embeddings to ELECTOR and FlexiBERT 2.0, respectively.

We now discuss the dynamic inference and training technique, DynaProp, that prunes activations and gradients to skip 
ineffectual operations. We then present the ELECTOR simulator and the accelerator design choices it supports. Finally, 
we describe the TransCODE pipeline that implements co-design and obtains the best-performing transformer-accelerator pair.

\newcolumntype{A}{ >{$} r <{$} @{} >{${}} l <{$}} 
\begin{table}[]
\caption{Forward and backward pass operations for matrix multiplication and 1D convolution.}
\centering
\begin{tabular*}{\linewidth}{
  @{\hspace{\tabcolsep}\extracolsep{\fill}}
  l@{\hskip 0.8in}A
  @{\hspace{\tabcolsep}}
}
\toprule
\multicolumn{2}{l}{\textbf{Matrix Multiplication}} \\ \midrule
Forward Pass & \mathbf{X}_i &= f_i(\mathbf{W}_i \mathbf{X}_{i-1}) \\ [2mm]
Backward Pass & \delta_i &= \mathbf{W}_{i+1}^\intercal \delta_{i+1} \cdot f'(\mathbf{W}_i \mathbf{X}_{i-1}) \\ [2mm]
Weight Update & \nabla_{\mathbf{W}_i} \mathcal{L} &= \delta_i \mathbf{X}_{i-1}^\intercal \\ [2mm]
 & \mathbf{W}_i &= \mathbf{W}_i - \alpha \cdot \nabla_{\mathbf{W}_i} \mathcal{L} \\ \midrule
\multicolumn{2}{l}{\textbf{1D Convolution}} \\ \midrule
Forward Pass & \mathbf{X}_i &= \mathbf{w}_i \ast \mathbf{X}_{i-1} \\ [2mm]
Backward Pass & \delta_i &= \nabla_{\mathbf{W}_i} \mathcal{L} \\ [2mm]
Weight Update & \nabla_{\mathbf{W}_i} \mathcal{L} &= \delta_i \ast \overleftarrow{\mathbf{X}_{i-1}} \\  [2mm]
 & \mathbf{W}_i &= \mathbf{W}_i - \alpha \cdot \nabla_{\mathbf{W}_i} \mathcal{L} \\ \bottomrule
\end{tabular*}
\label{tbl:backprop}
\end{table}

\subsection{Dynamic Inference and Training}
\label{sec:dynaprop}

DynaTran~\cite{acceltran} is a low-overhead dynamic inference method that quickly prunes ineffectual weight and 
activation values at runtime. However, it only targets transformer inference and not training. We propose DynaProp 
that induces sparsity in weights and activations at runtime (during inference) and gradients (during training). 
DynaProp takes an input matrix, which is either a weight matrix (loaded from memory), an activation matrix (obtained from 
previous MAC operations), or a gradient matrix (formed while backpropagating gradients). It then prunes values with 
a magnitude less than a given threshold $\tau$ (i.e., it forces them to zero). Mathematically, we prune an input 
matrix $\mathbf{M} \in \mathbb{R}^{m \times n}$ to $\mathbf{M}^\text{P}$ as follows:
\begin{equation*}
    \mathbf{M}^\text{P}_{ij} = 
    \begin{cases*} 
        \mathbf{M}_{ij} & if $ \lvert \mathbf{M}_{ij} \rvert \ge \tau $  \\
        0 & if $ \lvert \mathbf{M}_{ij} \rvert < \tau $
    \end{cases*}
\end{equation*}

We implement each such comparison in parallel, thus requiring only one clock cycle for the pruning process. We define the pruning ratio 
(or level of sparsity) of the output matrix as:
\begin{equation*}
    \rho(\mathbf{M}^\text{P}) = \frac{\sum_{x \in \mathbf{M}^\text{P}} \delta_{x, 0}}{m \times n}
\end{equation*} where $\delta$ is the Kronecker delta function. We profile the resultant sparsity in weights, activations, and gradients for different transformer models on diverse applications to obtain a desired $\rho$. ELECTOR stores these curves in memory. For the desired values of $\rho$, we determine the corresponding $\tau$ at runtime through a simple look-up operation. We present such curves in Section~\ref{sec:results_dynaprop}.

Table~\ref{tbl:backprop} shows the operations underlying the forward and backward pass for matrix multiplication and
one-dimensional (1D) convolution, respectively. The table shows that training requires the same operation types (as
inference) and thus mandates identical hardware, although with a separate dataflow. We also observe that the number
of backward pass and weight update operations (executed during training) is more than the number of those for the 
forward pass (executed during inference). This shows that training is much more computationally expensive than 
inference, involving more activations and gradients that the accelerator needs to account for. DynaProp prunes each 
such matrix before it executes the respective operation in hardware. Thus, the accelerator skips ineffectual operations, 
improving latency and energy efficiency.

\textcolor{black}{Optimizers like Adam would require extra computation (e.g., the calculation of momentum and storage of previous
weights/gradients). These computations can easily be incorporated into the accelerators supported in the proposed design space. However, second-order gradients would add much more computational overhead. We leave the application of complex optimizers to future work.}

\subsection{The ELECTOR Framework}
\label{sec:elector}

\begin{figure}
    \centering
    \includegraphics[width=0.9\linewidth]{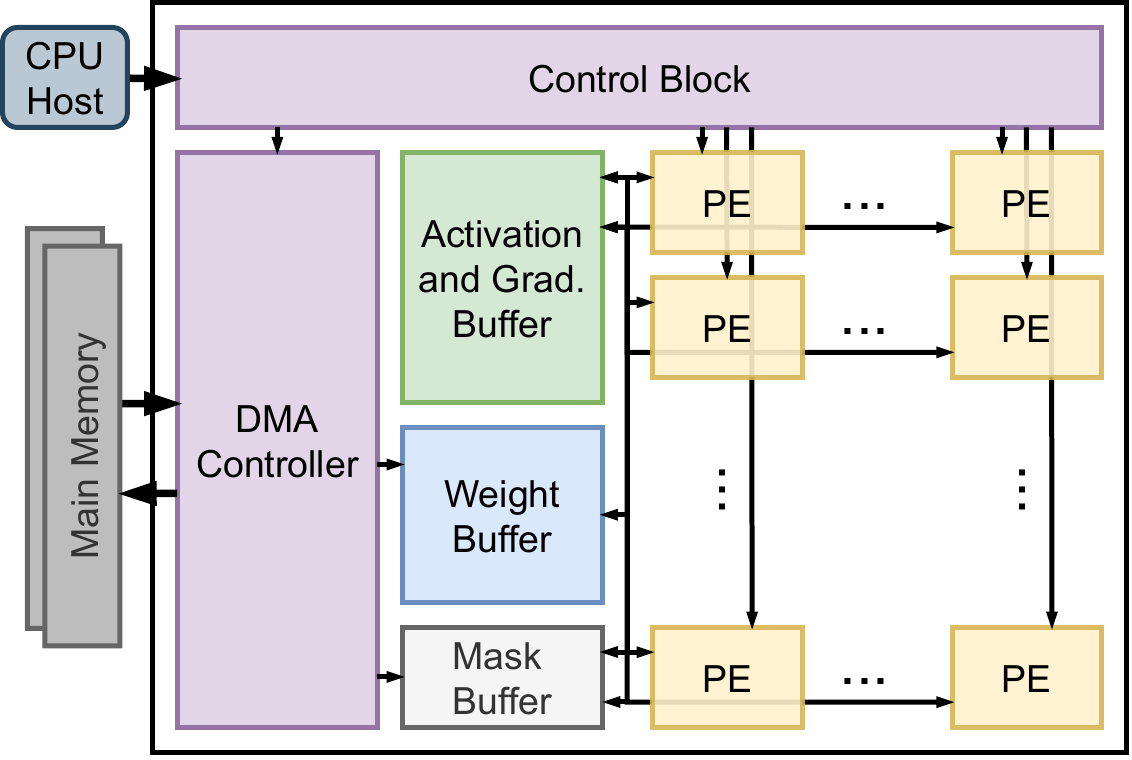}
    \caption{Organization of a typical accelerator in the ELECTOR design space.}
    \label{fig:elector_org}
\end{figure}

Accelerators in the ELECTOR design space take inspiration from previously proposed state-of-the-art accelerators,
including SPRING~\cite{spring} and AccelTran~\cite{acceltran}. We divide the overall accelerator architecture into
the accelerator tier and the (on-chip or off-chip) memory tier. Fig.~\ref{fig:elector_org} shows the organization of
the accelerator tier in the proposed architecture. The control block receives the instruction stream of the transformer model from the host CPU. The direct memory access (DMA) controller fetches the weights and embeddings from the main memory. \textcolor{black}{Thus, the PEs communicate with the on-chip buffers while the DMA controller transfers data between the buffers and the on-chip/off-chip main memory.} The activation-and-gradient buffer stores the activations and gradients formed during transformer evaluation. The weight buffer stores the transformer weights. ELECTOR stores all data in a compressed format (discussed in Section~\ref{sec:elector_modules}). Data compression relies on binary masks (stored in the mask buffer). The PEs employ the compressed data and 
associated masks to perform the main compute operations of any transformer in the FlexiBERT 2.0 design space.

\subsubsection{Hardware Modules}
\label{sec:elector_modules}

\begin{figure}
    \centering
    \includegraphics[width=0.9\linewidth]{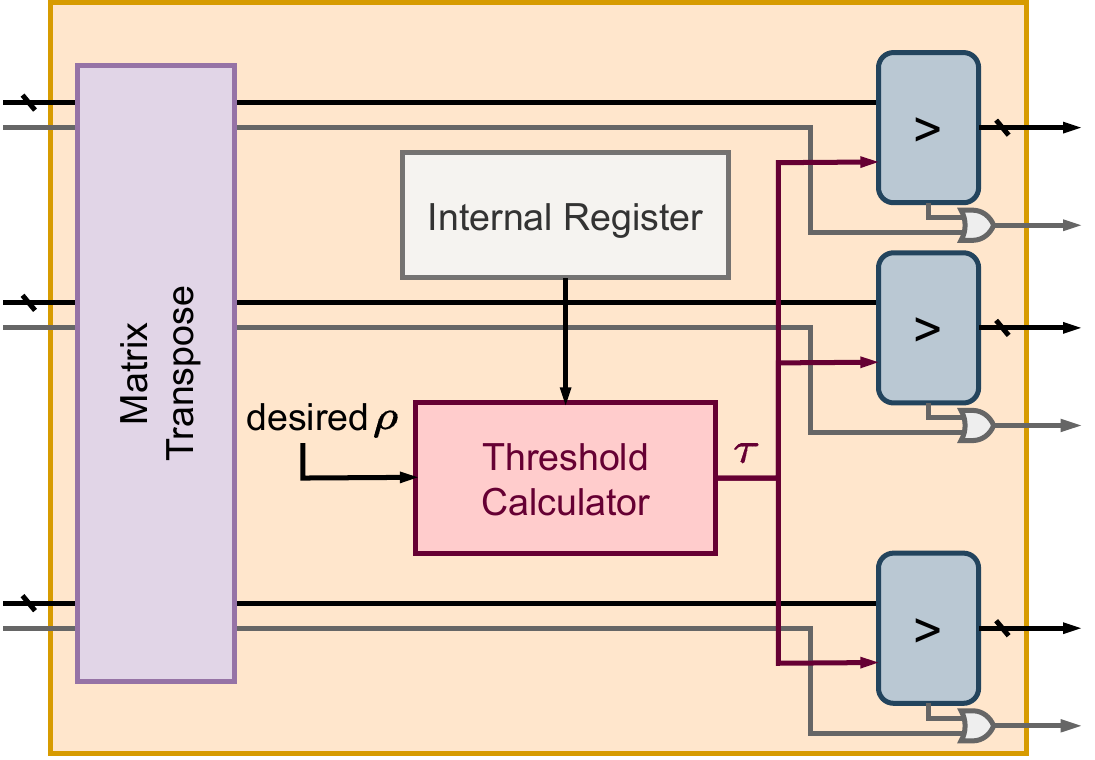}
    \caption{Implementation of the DynaProp module. The wires for mask bits are in grey.}
    \label{fig:dynaprop}
\end{figure}

We now describe various modules supported in the ELECTOR design space.
\begin{itemize}
    \item \myuline{Main Memory}: ELECTOR supports three memory types: an off-chip dynamic random access memory (DRAM) for scalable and economical deployments, an on-chip high-bandwidth memory (HBM) for memory-intensive edge/server applications, and an on-chip monolithic-3D resistive random access memory (RRAM). Monolithic-3D integration leverages monolithic inter-tier vias, allowing much higher density than traditional through-silicon-via-based 3D integration~\cite{miv}. This leaves much more logic space and permits high memory bandwidth, which are crucial for large transformer models in the FlexiBERT 2.0 design space.
    \item \myuline{Control Block}: The control block takes the transformer model as input. It then converts all
functions in the model into hardware-mappable operations (details in Section~\ref{sec:elector_mapper}) that it later
converts to \emph{tiled} operations. For instance, it converts the matrix multiplication operation \textbf{\texttt{O
= W$\times$A}} to multiple operations of the form \textbf{\texttt{O[b,i,j] = W[b,i,k]$\times$A[b,k,j]}}, where each tiled matrix $\in \mathbb{R}^{b \times x \times y}$, i.e., the tile size~\cite{acceltran}. The control block also assigns and schedules the tiled operations to different PEs~\cite{acceltran}.
\begin{figure}
        \centering
        \includegraphics[width=0.9\linewidth]{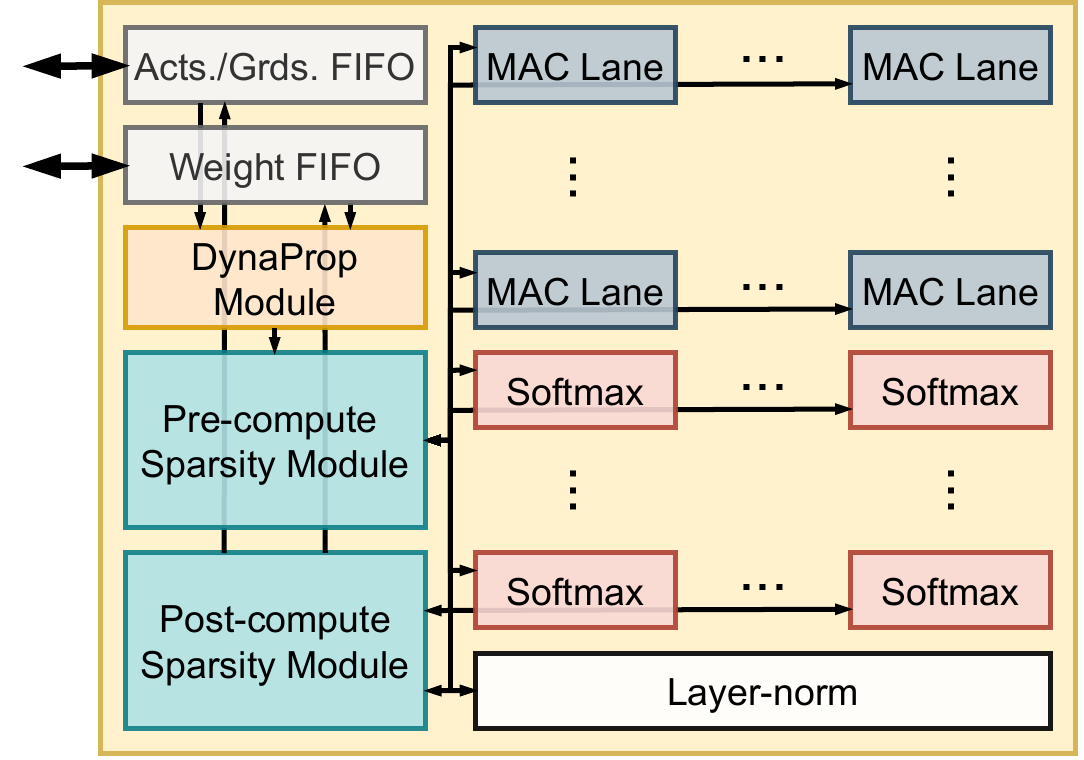}
        \caption{Internal components of a PE.}
        \label{fig:pe}
    \end{figure}
    \item \myuline{Processing Elements}: Fig.~\ref{fig:pe} shows the organization of a PE (the basic compute module 
of an accelerator) in the ELECTOR design space. The local registers of the PE store the compressed data. These are
the first-in-first-out (FIFO) registers for the activations (and gradients) and weights. The data then enter the
DynaProp module that induces sparsity based on the desired $\rho$. As explained in Section~\ref{sec:dynaprop}, this
module prunes the given activation/gradient/weight matrices based on a pre-calculated threshold $\tau$. The PE then
feeds the sparse data to the pre-compute sparsity module with the binary masks. These binary masks have the same
shape as the uncompressed data, where each binary bit in a mask depicts if the corresponding element in the original data vector is ineffectual or not. The pre-compute sparsity module converts the input data into a zero-free format based on the associated masks~\cite{acceltran}. The PE then forwards this zero-free data to the MAC lanes (for matrix multiplication), softmax modules (for softmax operation), or the layer-norm module (for layer-norm operation). The zero-free data eliminate any ineffectual computations in these modules. Finally, the post-compute sparsity module~\cite{acceltran} implements the inverse of this operation on the output activations before storing them in the corresponding FIFO register and, eventually, the main buffer. 
    \begin{itemize}
        \item The MAC lanes execute multiplication between two tiles in a parallelized manner. We store all
activation, gradient, and weight data in fixed-point format with $(\text{IL} + \text{FL})$ bits, denoting integer
length (IL) and fractional length (FL), respectively~\cite{acceltran}. Data first reach $M$ multipliers and then an adder tree with depth $\log_2 M$. The MAC lanes also include a ReLU and a GeLU~\cite{gelu} module for feed-forward operations.
        \item Fig.~\ref{fig:dynaprop} shows the DynaProp module that executes dynamic inference and training on the
transformer. It takes the input activation/gradient/weight matrix and prunes ineffectual values for efficient
evaluation. As explained in Section~\ref{sec:dynaprop}, we prune the values of the input matrix by comparing their
magnitude with a pre-determined threshold $\tau$. The DynaProp module implements this in parallel for the entire
tile. We first feed an input tile $\mathbf{M} \in \mathbb{R}^{b \times x \times y}$ to the matrix transpose block,
which carries out the transpose operation, if required. Mathematically, it outputs $\mathbf{M}^\intercal \in
\mathbb{R}^{b \times y \times x}$, transposing all matrices in the batch of size $b$. It then feeds the input tile
to $b \times x \times y$ comparators. The threshold calculator determines the required threshold using the desired
$\rho$ and the pre-profiled transfer functions for different transformer models on diverse applications (stored in
the internal register; more details in Section~\ref{sec:results_dynaprop}). If the output of the comparator is zero,
we set the corresponding mask bit to 1. Here, we represent the lines carrying mask information in grey and those carrying 
activation/gradient/weight information in black.
        \item For all other hardware modules, we use the proposed implementation of AccelTran~\cite{acceltran}.
However, we expand the operation executability of all modules (e.g., support for different tile sizes in the softmax
module), as explained in Section~\ref{sec:elector_space}.
    \end{itemize}
\end{itemize}

The optimal selection of the number of PEs, buffer sizes, and other design choices results in the highest possible resource utilization while minimizing the number of compute/memory stalls (when we do not execute either a compute operation or a memory fetch operation). Hence, determining the best accelerator hyperparameters is essential for energy-efficient designs 
with a low chip area and high throughput.

\subsubsection{The Transformer Mapper}
\label{sec:elector_mapper}

The FlexiBERT 2.0~\cite{edgetran} design space supports various operation types. We describe each operation next.

\begin{itemize}
    \item \myuline{Self-attention}: The self-attention (SA) operation finds how much one token \emph{attends} to another token. For an output attention head $\mathbf{H}_i \in \mathbb{R}^{N_T \times d_{out}}$ with query $\mathbf{Q}_i \in \mathbb{R}^{N_T \times h/n}$, key $\mathbf{K}_i \in \mathbb{R}^{N_T \times h/n}$, and value $\mathbf{V}_i \in \mathbb{R}^{N_T \times h/n}$ matrices~\cite{flexibert}:
    \begin{equation*}
            \mathbf{H}_i = \text{softmax}\left(\text{SA}\right) \mathbf{V}_i \mathbf{W}^o_i
        \end{equation*}
    where $N_T$ is the input sequence length, $h$ is the hidden dimension of the encoder layer, and $n$ is the number of heads. The SA operation has two sub-types:
    \begin{itemize}
        \item The scaled dot-product (SDP) attention~\cite{vaswani} is the de-facto standard operation in traditional transformer architectures. Mathematically, 
        \begin{equation*}
            \text{SA}_\text{SDP} := \frac{\mathbf{Q}_i \mathbf{K}_i^\intercal}{\sqrt{h}}.
        \end{equation*}
        \item The weighted multiplicative attention (WMA)~\cite{wma} involves a trainable weight matrix $\mathbf{W}_a \in \mathbb{R}^{h/n \times h/n}$ such that
        \begin{equation*}
            \text{SA}_\text{WMA} := \frac{\mathbf{Q}_i \mathbf{W}_a \mathbf{K}_i^\intercal}{\sqrt{h}}.
        \end{equation*}
        The mapper converts the self-attention operation into various MAC and softmax operations that the corresponding hardware modules can execute in the accelerator.
    \end{itemize}
    \item \myuline{Linear Transform}: As the name suggests, this operation implements a linear transform (LT) on the input sequence. The FlexiBERT 2.0 design space supports two sub-types:
    \begin{itemize}
        \item The discrete Fourier transform (DFT) that we implement in hardware using the corresponding Vandermonde matrix $\mathbf{V}_\text{DFT} \in \mathbb{R}^{N_T \times N_T}$ for the roots of unity (also called the DFT matrix)~\cite{dft_book} such that
        \begin{equation*}
            \text{LT}_\text{DFT} := \mathbf{V}_\text{DFT} \ \mathbf{H}
        \end{equation*}
        where $\mathbf{H} \in \mathbb{R}^{N_T \times d_{in}}$ represents a matrix for the input hidden states.
        \item The discrete cosine transform (DCT) that we again implement using an equivalent Vandermonde matrix $\mathbf{V}_\text{DCT} \in \mathbb{R}^{N_T \times N_T}$ such that
        \begin{equation*}
            \text{LT}_\text{DCT} := \mathbf{V}_\text{DCT} \ \mathbf{H}
        \end{equation*}
    \end{itemize}

We store the $\mathbf{V}_\text{DFT}$ and $\mathbf{V}_\text{DCT}$ matrices in the buffer for subsequent use while
executing the above operations. Although these operations are slower than the fast Fourier transform
(FFT)~\cite{fft} and the fast cosine transform (FCT)~\cite{fct}, respectively, sparsification of the input matrices
results in a low overall execution time. Furthermore, converting these operations to MAC operations enables the
reuse of the MAC lanes, thus not requiring special hardware modules for the LT operation. Nevertheless, these methods 
(FFT and FCT) may lead to high gains for transformer models that support long sequences~\cite{longformer}, due to 
their $\mathcal{O}(N \log N)$ complexity. We leave their hardware implementation to future work.
    \item \myuline{Dynamic-span-based Convolution}: The dynamic-span-based convolution (DSC) operation implements a 1D convolution over the input. Mathematically, 
    \begin{equation*}
        \text{DSC}_k := \mathbf{w}_k \ast \mathbf{H}
    \end{equation*}
    where $\mathbf{w}_k$ is the convolution kernel of length $k$. To implement this operation in hardware, we convert the convolution operation into an equivalent matrix multiplication operation. In other words, we convert the convolutional kernel to a sparse matrix that we multiply with the input. We tweak the MAC lane module to incorporate this conversion.
\end{itemize}

Now that the mapper has converted the operations in the FlexiBERT 2.0 design space to hardware-implementable formats, the control block tiles, schedules, and assigns these mapped operations to the accelerator for transformer evaluation.

\subsubsection{Design Space}
\label{sec:elector_space}

ELECTOR supports various accelerators in its design space. It allows adaptation of many design decisions in an
ASIC-based accelerator. We describe these tunable hyperparameters next.

\begin{table}[]
\centering
\caption{Hyperparameters supported in the ELECTOR design space.}
\begin{tabular}{@{}ll@{}}
\toprule
\textbf{Hyperparameter} & \textbf{Permissible values} \\ \midrule
Batch tile size & 1, 4 \\
Spatial tile size & 8, 16, 32 \\
Activation function & ReLU, GeLU \\
\#PEs & 64, 128, 256, 512, 1024 \\
\#MAC lanes per PE & 8, 16, 32, 64, 128 \\
\#MACs per lane & 1, 16 \\
\#Softmax modules per PE & 2, 4, 8, 16, 32, 64 \\
Batch size & 4, 16, 32 \\
Act./grad. buffer size (MB) & 4, 8, 16, 32, 64 \\
Weight buffer size (MB) & 8, 16, 32, 64, 128 \\
Mask buffer size (MB) & 1, 2, 4, 8, 16 \\
\begin{tabular}[c]{@{}l@{}}Main memory configuration \\ {[}banks, ranks, channels{]} \end{tabular} & \begin{tabular}[c]{@{}ll@{}}RRAM: & [16, 2, 2], [8, 2, 4], [4, 2, 8],  \\ & [2, 2, 16], [32, 2, 1], [1, 2, 32]\\ DRAM: & [16, 2, 2], [8, 2, 4], [32, 2, 1], \\ & [16, 4, 1] \\ HBM: & [32, 1, 4]\end{tabular} \\ \bottomrule
\end{tabular}
\label{tbl:elector_design_space}
\end{table}

\begin{itemize}
    \item \myuline{Batch Tile Size}: This is the size of a tile along the batch. Mathematically, a tile $\mathbf{M} \in \mathbb{R}^{b \times x \times y}$ has the batch tile size $b$.
    \item \myuline{Spatial Tile Size}: This is the size of a tile orthogonal to the batch dimension. In the above example, $x = y$ is the spatial tile size (we assume square matrices for the tiles). A higher tile size (either $b$ or $x$/$y$) would imply that each hardware module (MAC lane, softmax module, or layer-norm module) could execute more 
operations in parallel since the module evaluates a larger tile. This enables latency reduction at the cost of higher dynamic power.
    \item \myuline{Activation Function}: Transformer evaluation uses a nonlinear function following a feed-forward 
operation. We support two functions: ReLU and GeLU~\cite{gelu}. This is in accordance with the FlexiBERT 2.0 design space.
    \item \myuline{Number of PEs}: The number of PEs in the accelerator.
    \item \myuline{Number of MAC Lanes per PE}: The number of MAC lanes in each PE of the accelerator. We keep the number 
of MAC lanes constant for every PE.
    \item \myuline{Number of MACs per Lane}: The number of MAC units per MAC lane. Again, this is constant across 
all MAC units.
    \item \myuline{Number of Softmax Modules per PE}: The number of softmax modules in each PE. Every PE has only one layer-norm module. Therefore, the number of MAC lanes and softmax modules in each PE determines the net ratio of the number of MAC lanes, softmax modules, and layer-norm modules in an accelerator. One can tune this ratio based on the corresponding proportion of these operations in evaluating the selected transformer.
    \item \myuline{Batch Size}: The batch size for transformer evaluation. More compute resources and high bandwidth memory enable a larger batch, reducing evaluation latency.
    \item \myuline{Activation and Gradient Buffer Size}: The size of the activation/gradient buffer. Training 
requires more activation matrices than inference. It also has gradient matrices, requiring a larger buffer size.
    \item \myuline{Weight Buffer Size}: The size of the weight buffer. A larger transformer model 
requires a larger weight buffer.
    \item \myuline{Mask Buffer Size}: The size of the mask buffer that stores the binary masks for the zero-free format~\cite{acceltran} used in the accelerators in ELECTOR.
\end{itemize}

Table~\ref{tbl:elector_design_space} summarizes the possible design choices for accelerators in the ELECTOR design space. 
The possible memory configurations include the memory type (RRAM, DRAM, and HBM) along with the banks, ranks, and channels. 

\subsubsection{Accelerator Embeddings}
\label{sec:elector_embeddings}

We now describe how we convert the selected accelerator configuration (a sample from Table~\ref{tbl:elector_design_space}) 
to an embedding for surrogate modeling. We generate a 12-dimensional embedding ($e$) for a selected accelerator 
configuration as follows:

\begin{itemize}
    \item $e_1$ denotes the batch tile size, i.e., $e_1 = b$.
    \item $e_2$ and $e_3$ correspond to the spatial tile sizes, i.e., $e_2 = x$, $e_3 = y$. For the targeted design space, $e_2 = e_3$.
    \item $e_4$ denotes the number of PEs.
    \item $e_5$ denotes the number of MAC lanes per PE.
    \item $e_6$ denotes the number of MACs per lane.
    \item $e_7$ denotes the number of softmax modules in each PE.
    \item $e_8$ denotes the selected batch size for model evaluation.
    \item $e_9$, $e_{10}$, and $e_{11}$ denote the activation/gradient, weight, and mask buffer sizes, respectively, in 
MBs.
    \item $e_{12}$ denotes the index of possible memory configurations in Table~\ref{tbl:elector_design_space}, thus 
ranges from 1 to 11.
\end{itemize}

We use these generated embeddings to train the TransCODE surrogate model, which also outputs the subsequent query as an 
accelerator embedding.

\subsubsection{Simulation Flow}
\label{sec:elector_simulation}

\begin{figure}
    \centering
    \includegraphics[width=\linewidth]{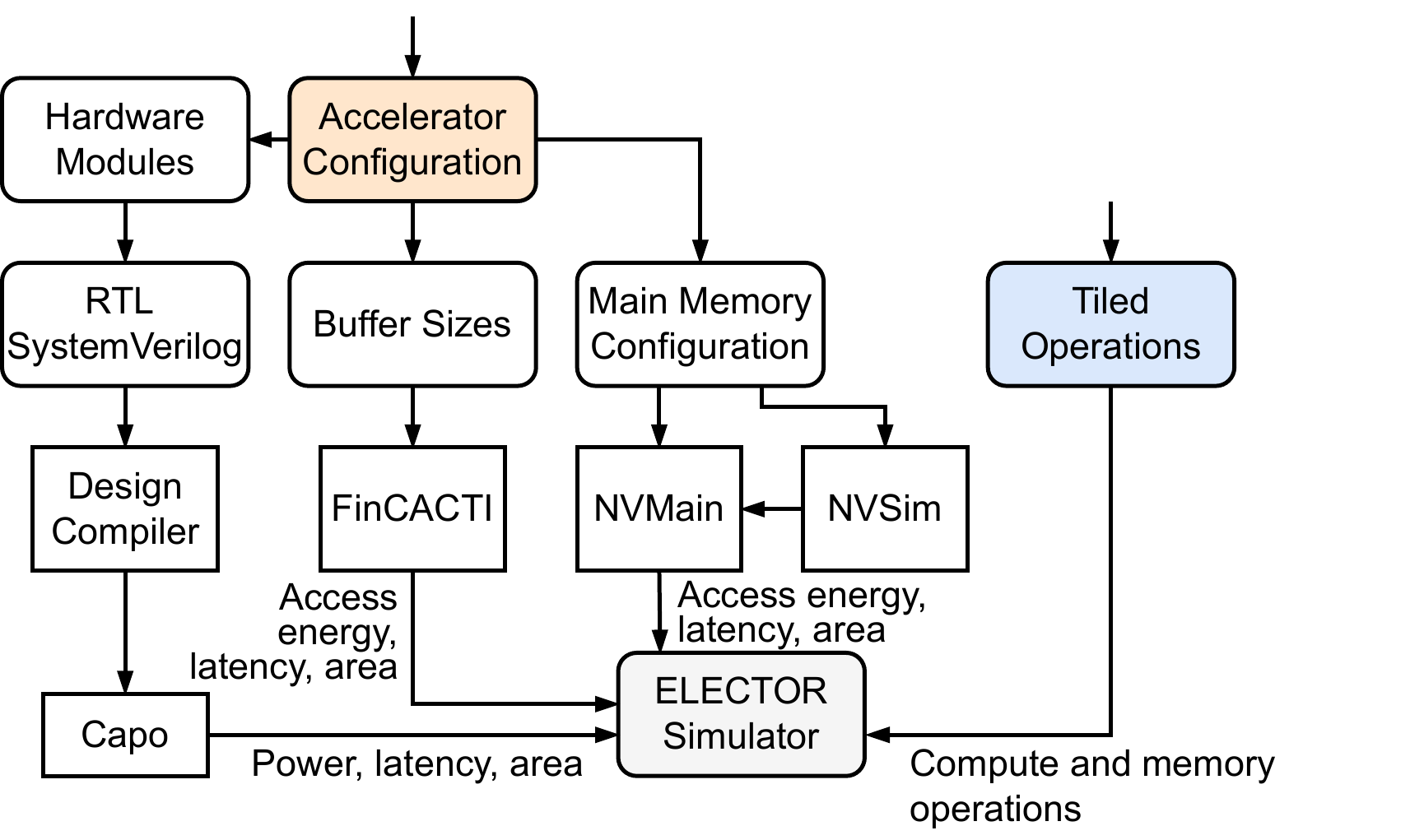}
    \caption{Flow of simulation in ELECTOR.}
    \label{fig:simulator}
\end{figure}

Fig.~\ref{fig:simulator} shows the simulation flow for evaluating an input accelerator configuration and tiled 
operations (obtained after mapping and tiling the input transformer) in ELECTOR. We first select the compute 
modules (including the tile size for parallel operation), buffer sizes, and main memory configuration. Next, we 
implement different hardware modules discussed in Section~\ref{sec:elector_modules} at the register-transfer level 
(RTL) using SystemVerilog. We use Design Compiler~\cite{dc} to synthesize the RTL design based on a 14nm FinFET 
technology library~\cite{14nm}. Capo~\cite{capo}, an open-source floorplacer, performs floorplanning. 
FinCACTI~\cite{fincacti}, a cache modeling tool for deeply-scaled FinFETs, models the on-chip buffers. 
NVSim~\cite{nvsim} and NVMain~\cite{nvmain} model the main memory (either the off-chip DRAM or on-chip HBM/RRAM). 
ELECTOR then plugs the synthesized results into a Python-based cycle-accurate simulator. Finally, the control block 
segregates the tiled operations into compute and memory operations for separate execution pipelines~\cite{acceltran}.

\subsection{TransCODE}
\label{sec:transcode}

We use BOSHCODE to obtain the best-performing transformer-accelerator pair. BOSHCODE takes as input the accelerator and 
transformer embeddings and outputs the performance measure to be estimated. For the transformer embeddings, we use the 
embeddings used in FlexiBERT 2.0~\cite{edgetran} as opposed to the \texttt{Transformer2vec} encodings~\cite{flexibert} 
since they are fast and efficient. This is critical for exploring the vast FlexiBERT 2.0 design space efficiently. For 
the accelerator embeddings, we use the embeddings from the accelerator configuration discussed in 
Section~\ref{sec:elector_embeddings}. We define the output performance measure as follows:
\begin{align*}
\label{eqn:perf_metric}
\begin{split}
    \text{Performance} &= \alpha \times (1 - \text{Latency}) + \beta \times (1 - \text{Area}) \\
    &+ \gamma \times (1 - \text{Dynamic Energy}) \\
    &+ \delta \times (1 - \text{Leakage Energy}) + \epsilon \times \text{Accuracy}
\end{split}
\end{align*}
where $\alpha + \beta + \gamma + \delta + \epsilon = 1$ are hyperparameters. We normalize the values of the individual 
performance measures with respect to their maximum values (hence, these values reside in the $[0,1]$ interval). Thus, 
for edge applications where the power envelope of devices is highly restricted, users can set the hyperparameters 
$\gamma$ and $\delta$ high. On the other hand, for server-side deployments, where accuracy is of utmost importance, 
one can set $\epsilon$ high.

TransCODE needs five performance values for the queried transformer-accelerator pair: latency, area, dynamic energy, 
leakage energy, and model accuracy. To obtain the first four performance values, we leverage the ELECTOR simulator. To
obtain the transformer model accuracy, we employ the FlexiBERT 2.0 surrogate model, which outputs the GLUE 
score~\cite{glue}.

\begin{table}[t]
\caption{Hyperparameter ranges in FlexiBERT 2.0 design space~\cite{edgetran}. Super-script ($j$) depicts the value for layer $j$.}
\vskip 0.1in
    \centering
    \resizebox{\columnwidth}{!}{
    \begin{tabular}{@{}ll@{}}
    \toprule
        \textbf{Design Element} & \textbf{Allowed Values} \\
        \midrule 
        Number of encoder layers ($l$) & $\{2,4,6,8,10,12\}$\\
        Type of attention operation used ($o^j$) & $\{\text{SA},\text{LT},\text{DSC}\}$\\
        Number of operation heads ($n^j$) & $\{2,4,8,12\}$ \\
        Hidden size ($h^j$) & $\{128,256\}$\\
        Feed-forward dimension ($f^j$) & $\{256,512,1024,2048,3072,4096\}$\\
        Number of feed-forward stacks & $\{1,2,3\}$\\
        Operation parameters ($p^j$): & \\
       \hspace{3mm} if $o^j = \text{SA}$ & Self-attention type: $\{\text{SDP},\text{WMA}\}$ \\ 
       \hspace{3mm} else if $o^j = \text{LT}$ & Linear transform type: $\{\text{DFT}, \text{DCT}\}$ \\
       \hspace{3mm} else if $o^j = \text{DSC}$ & Convolution kernel size: $\{5,9\}$ \tabularnewline
       \bottomrule
    \end{tabular}}
    \label{tbl:flexibert_2_design_space}
\end{table}

\section{Experimental Setup}
\label{sec:exp_setup}

In this section, we present the setup behind various experiments we performed, along with the baselines considered 
for comparison.

\subsection{Evaluation Models and Datasets}

To test the efficacy of the DynaProp method, we evaluate transformer models in the FlexiBERT 2.0 design space. 
Table~\ref{tbl:flexibert_2_design_space} shows the hyperparameter ranges supported by the FlexiBERT 2.0 design 
space~\cite{edgetran}. Evidently, shallow models (e.g., with two encoder layers) incur lower latency relative to 
deep models (e.g., with 12 encoder layers)~\cite{edgetran, hat_mit}. Moreover, wide models (e.g., with 12 attention heads) 
require more compute resources to enable higher parallelization than narrow ones (e.g., with two attention heads). 
Further, different attention-head types have different latencies and energy consumption characteristics. Hence, 
there is a need for optimized dataflows when executing such heterogeneous architectures.

We test the models on representative natural language understanding tasks under the GLUE benchmark~\cite{glue}. The 
included tasks are: SST-2~\cite{sst2}, MNLI~\cite{mnli}, QQP, QNLI, MRPC~\cite{qqp_qnli_mrpc}, CoLA~\cite{cola}, 
STS-B~\cite{stsb}, RTE~\cite{rte}, and WNLI~\cite{wnli}. The surrogate model trained on the FlexiBERT 2.0 design 
space~\cite{edgetran} reports the overall GLUE score. We show the training sizes and used metrics in 
Table~\ref{tbl:glue_summary}. The GLUE score represents average performance across all the tasks.

While running DynaProp, we target activation, weight, and gradient sparsity. Weight sparsity is static and depends on 
pruning performed during model pre-training or finetuning~\cite{movement_pruning}. Activation and gradient sparsity 
change for every input sequence -- we report their averages over the entire validation set.

\subsection{The ELECTOR Design Space}

Table~\ref{tbl:elector_design_space} summarizes the ELECTOR design space. Taking into account all the possible 
combinations presented in this table, ELECTOR supports 14,850,000 accelerators in its design space. This space
includes accelerators meant for resource-constrained edge applications as well as those relevant to high-energy server 
settings that require high throughput. In addition, ELECTOR allows different memory configurations to support diverse 
user requirements, from high-bandwidth monolithic-3D RRAM to economic off-chip DRAM.

\begin{table}[]
\caption{Data statistics of datasets in the GLUE benchmark.}
\centering
\begin{tabular}{@{}l@{\hskip 0.8in}l@{\hskip 0.8in}l@{}}
\toprule
Task & Training Size & Metric \\ \midrule
SST-2 & 67K & Accuracy \\
MNLI & 393K & Accuracy \\
QQP & 364K & Accuracy \\
QNLI & 105K & Accuracy \\
MRPC & 3.7K & Accuracy \\
CoLA & 8.5K & Matthew's Correlation \\
STS-B & 7K & Spearman Correlation \\
RTE & 2.5K & Accuracy \\
WNLI & 634 & Accuracy \\ \bottomrule
\end{tabular}
\label{tbl:glue_summary}
\end{table}

\subsection{Co-design Pipeline}

To run BOSHCODE, we use the following parameter values to obtain the net performance measure: $\alpha = 0.1$, 
$\beta = 0.1$, $\gamma = 0.2$, $\delta = 0.1$, and $\epsilon = 0.5$ (see Section~\ref{sec:transcode}). We leverage the 
network and hyperparameters used in EdgeTran~\cite{edgetran} for co-design. The BOSHCODE model takes $x_\text{TXF}$ and 
$x_\text{ACC}$ as input and outputs the predicted performance measure. Here, $x_\text{TXF}$ and $x_\text{ACC}$ correspond 
to the FlexiBERT 2.0 and ELECTOR embeddings, respectively. BOSHCODE then leverages gradient-based optimization using 
backpropagation to the input (GOBI)~\cite{tuli2021cosco} while freezing the model weights.

All input embeddings obtained using GOBI from the surrogate models may not be valid. For instance, $x_\text{ACC}$ 
should be well-defined (e.g., we allow the batch tile size, $b$, to only be 1 or 4). To add constraints to the 
optimization process, along with forcing the model to learn the performance only for valid input embeddings, we add a 
datapoint ($x_\text{TXF}$, $x_\text{ACC}$, $P_\text{MIN}$) to the dataset if either of the input embeddings is invalid or 
does not adhere to user-defined constraints. Another example of an input constraint could be that transformers with only 
up to six layers are allowed. $P_\text{MIN}$ has a low value, set to $-$1 for our experiments (where well-defined inputs 
would result in $P$ to lie in the [0,1] range).

\subsection{Evaluation Baselines}

We compare our experimental results with previously proposed transformer-accelerator pairs. The baseline accelerators 
include SpAtten and AccelTran, hand-designed for a specific transformer architecture. SpAtten implements HW-NAS. For 
fair comparisons, we present an HW-NAS version of TransCODE in which we execute BOSHCODE while forcing gradients to the 
accelerator to zero, i.e., we only search for transformer models run on a given edge platform. We also include co-design 
baselines implemented on a set of FPGAs~\cite{peng_dac_22}.

\begin{figure}
    \centering
    \includegraphics[width=\linewidth]{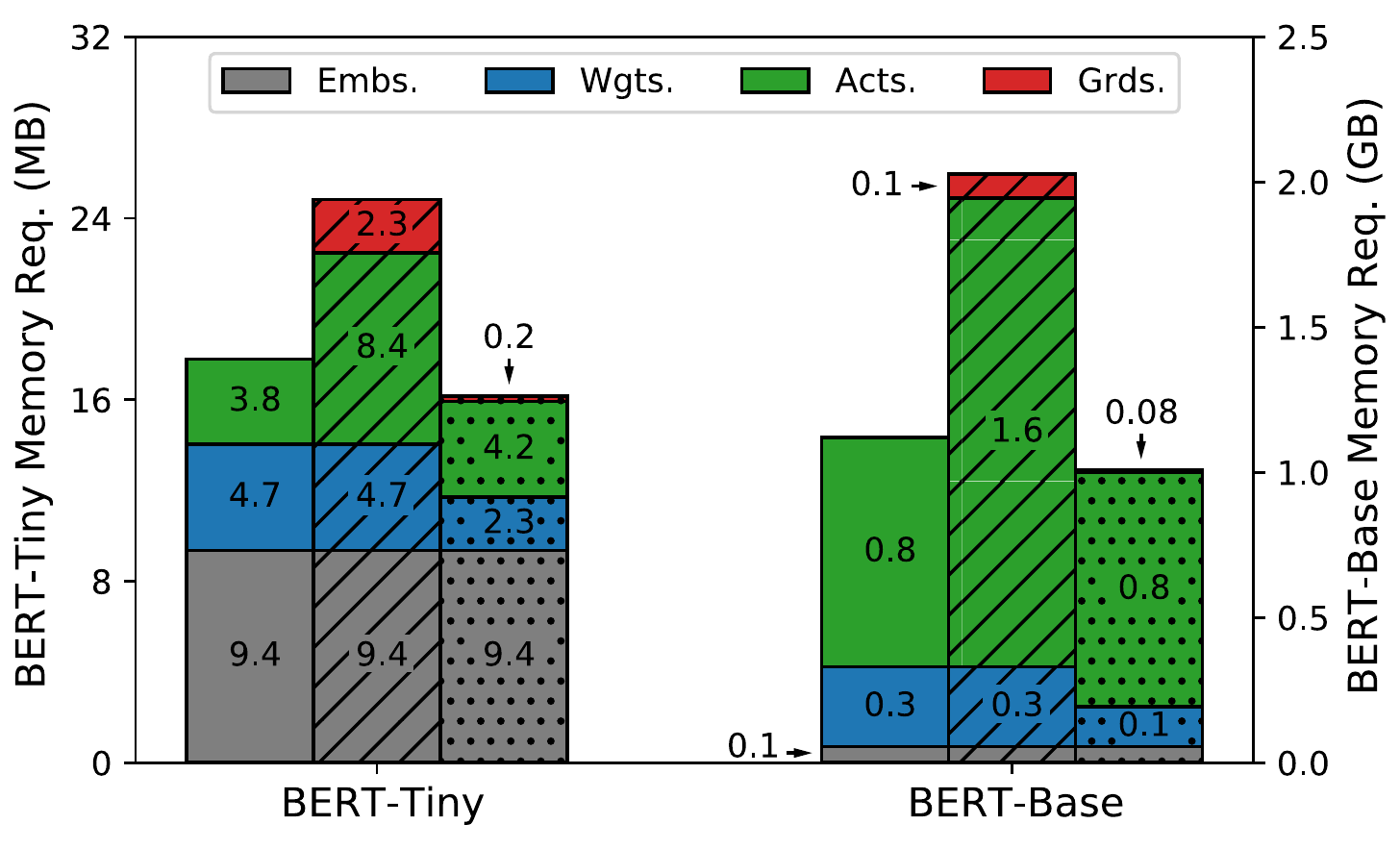}
    \caption{Breakdown of the total memory required while running inference (solid bars), traditional training 
(hatched bars), and DynaProp training (dotted bars) for BERT-Tiny and BERT-Base.}
    \label{fig:dynaprop_mem_req}
\end{figure}

\section{Experimental Results}
\label{sec:results}

In this section, we present the experimental results and comparisons of the TransCODE framework with relevant baselines.

\subsection{Dynamic Pruning of Transformer Weights, Activations, and Gradients}
\label{sec:results_dynaprop}

Fig.~\ref{fig:dynaprop_mem_req} shows a breakdown of the memory required for three evaluation modes: inference, 
traditional training, and DynaProp training, for BERT-Tiny and BERT-Base. The evaluation mode does not affect the 
memory usage for the token and position embeddings. Moreover, inference and training require the same memory size for 
transformer model weights. However, training generates gradients and also more activation operations. Here, the 
$\delta$'s described in Table~\ref{tbl:backprop} define the gradient memory consumption. For BERT-Tiny and BERT-Base, 
training requires 2.8$\times$ and 2.1$\times$ more memory (for activations and gradients), respectively. However, the 
buffer can be smaller since it only stores the activations or gradients required by the PEs at a given time. Finally, 
we show the memory required for DynaProp training. We configure DynaProp to induce 50\% sparsity in weights and 
activations (resulting in no loss in accuracy~\cite{acceltran}) and 90\% sparsity in the gradients (marginal accuracy 
loss, as shown below). DynaProp thus requires 1.5$\times$ and 1.9$\times$ smaller memory for BERT-Tiny and BERT-Base, 
respectively, while running training. This results in a smaller main memory, smaller buffers, and fewer MAC
operations, thus leading to improved throughput.

\begin{figure}
    \centering
    \includegraphics[width=\linewidth]{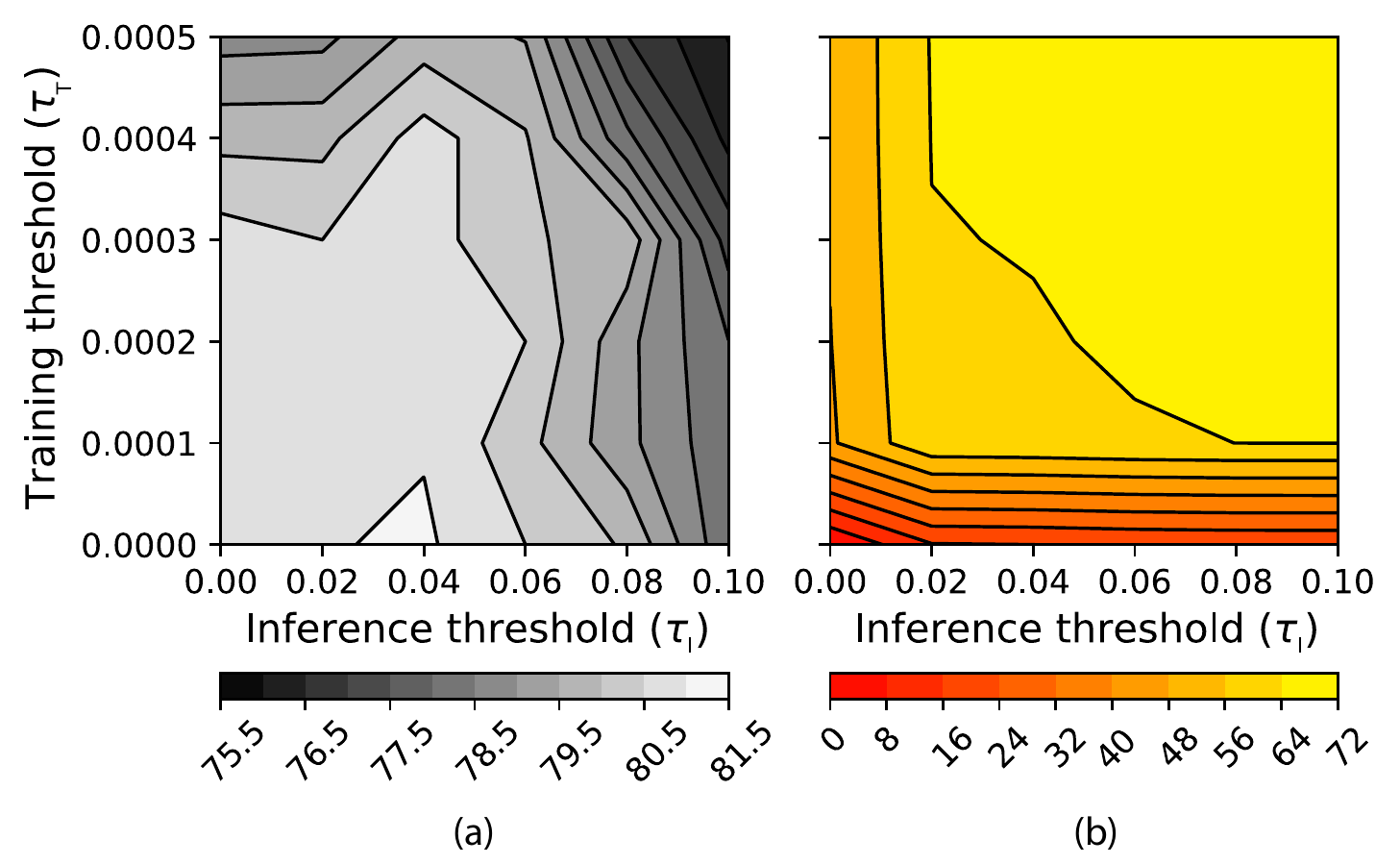}
    \caption{Effect of training ($\tau_\text{T}$) and inference ($\tau_\text{I}$) thresholds on (a) accuracy (\%) of
the SST-2 task and (b) averaged activation and gradient sparsity (\%) in BERT-Tiny.}
    \label{fig:tiny_acc_sp}
\end{figure}

To decouple and study the effects of pruning while running model inference and training, we execute DynaProp with two pruning thresholds: $\tau_\text{I}$ and $\tau_\text{T}$. It prunes activation and gradient matrices using $\tau_\text{I}$ for the forward pass and $\tau_\text{T}$ for the backward pass. It leverages movement pruning~\cite{movement_pruning} for transformer weights~\cite{acceltran}. Fig.~\ref{fig:tiny_acc_sp}(a) presents a contour plot showing the effect of these thresholds on accuracy for the BERT-Tiny model. As previously observed~\cite{energon, acceltran}, the accuracy first increases and then decreases as we increase $\tau_\text{I}$. However, accuracy monotonically decreases on increasing $\tau_\text{T}$. Fig.~\ref{fig:tiny_acc_sp}(b) shows the average between activation and gradient sparsities (or the net sparsity) when changing $\tau_\text{I}$ and $\tau_\text{T}$. The net sparsity increases as both $\tau_\text{I}$ and $\tau_\text{T}$ increase.

\begin{figure}
    \centering
    \includegraphics[width=\linewidth]{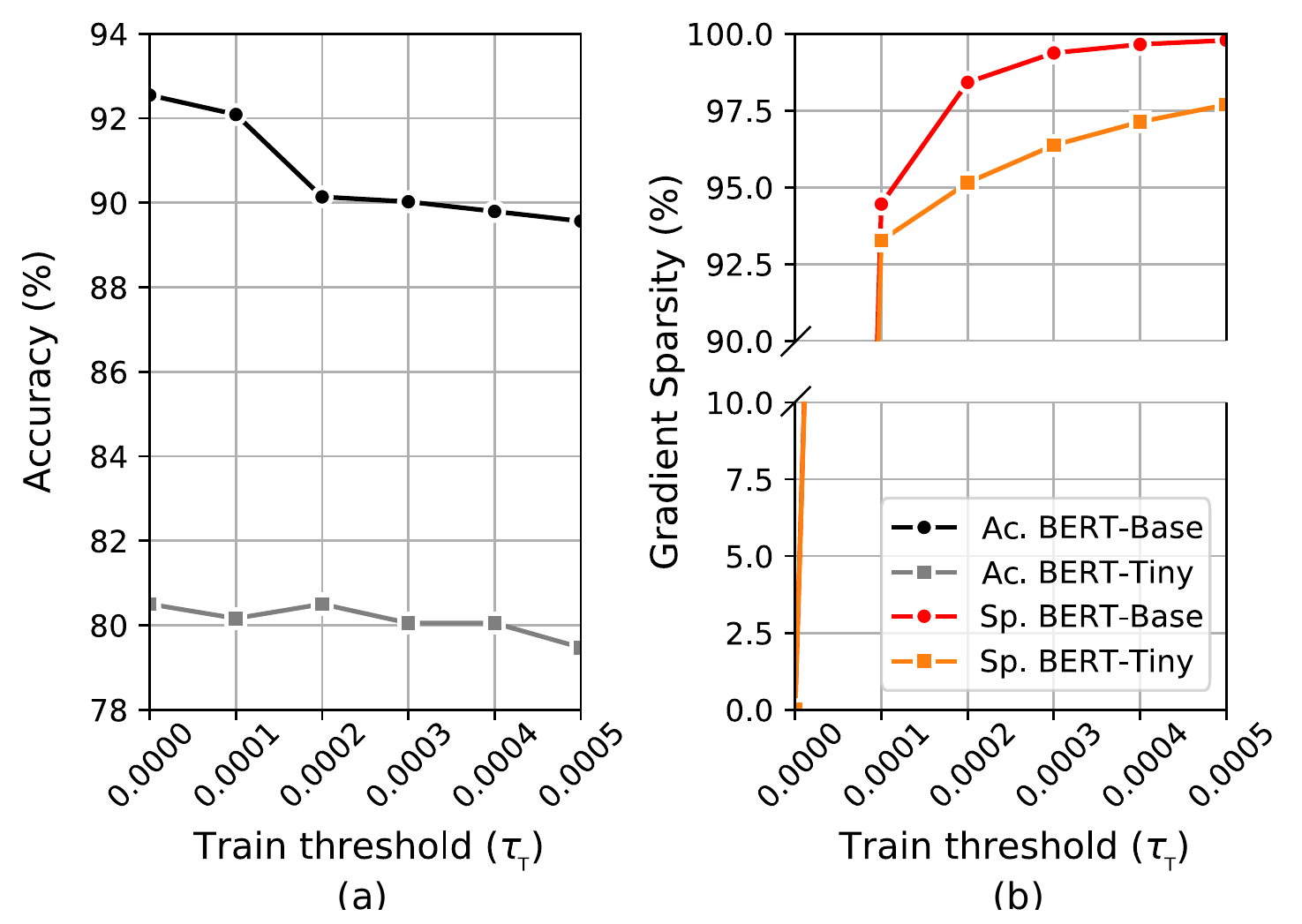}
    \caption{Effect of changing training threshold ($\tau_\text{T}$) on (a) accuracy of the SST-2 task and (b) sparsity 
in gradient matrices.}
    \label{fig:thresh_acc_sp}
\end{figure}

\begin{figure}
    \centering
    \includegraphics[width=0.95\linewidth]{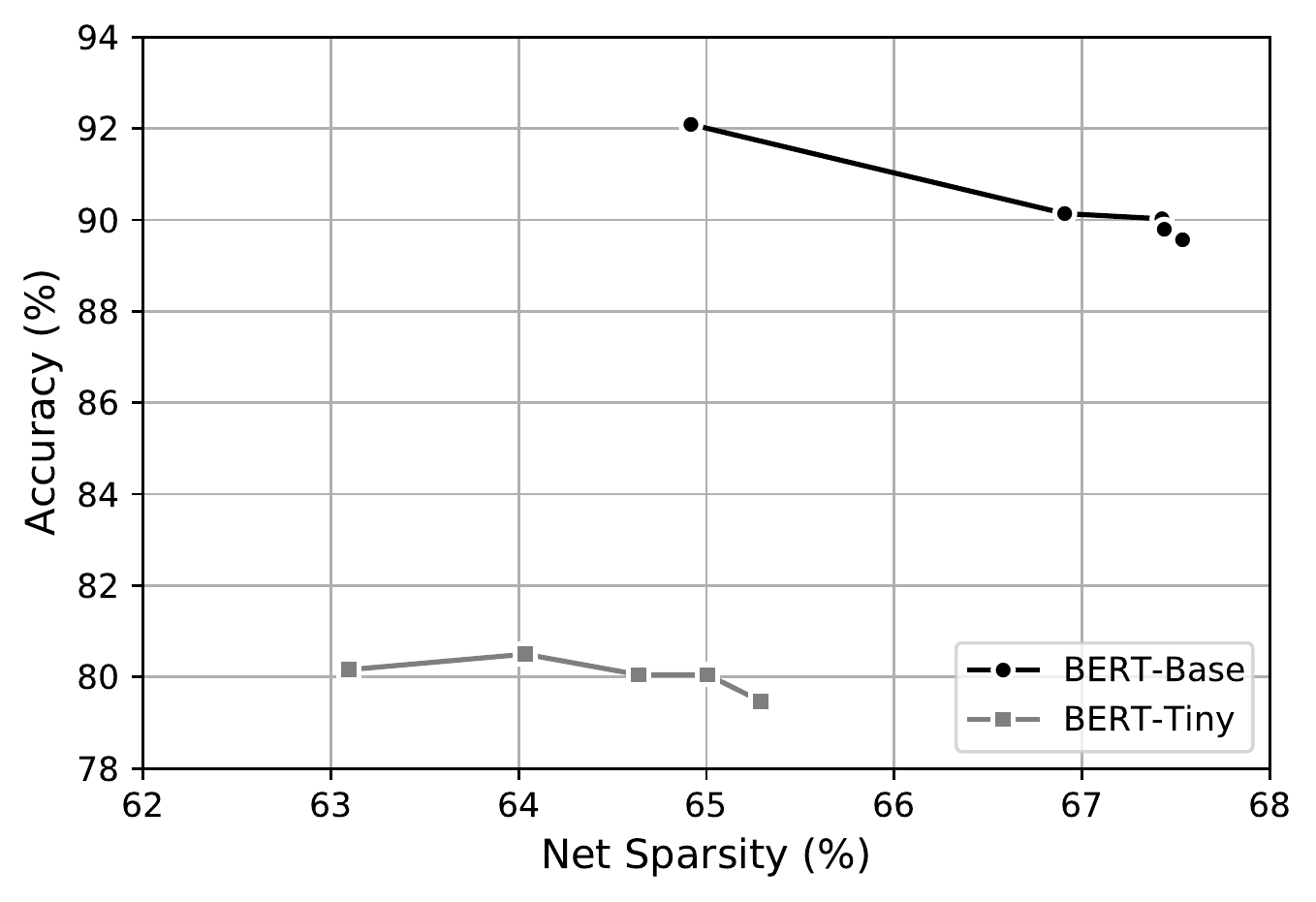}
    \caption{Accuracy of the SST-2 task plotted against net sparsity in activation and gradient matrices.}
    \label{fig:acc_sp}
\end{figure}

Fig.~\ref{fig:thresh_acc_sp}(a) shows accuracy plotted against $\tau_\text{T}$. Unlike pruning during inference
(using $\tau_\text{I}$), accuracy decreases under DynaProp as we increase the pruning threshold $\tau_\text{T}$. 
However, this loss in accuracy is a result of high gradient sparsity, as shown in Fig.~\ref{fig:thresh_acc_sp}(b). This 
enables ELECTOR to skip many ineffectual MAC operations, reducing energy consumption and latency. We achieve 90\% 
gradient sparsity when we set $\tau_\text{T}$ to 0.0001 with an accuracy loss of only 0.4\%. Fig.~\ref{fig:acc_sp} 
shows a plot of accuracy against net sparsity. Again, we define \emph{net sparsity} as the average of the 
activation and gradient sparsities (weight sparsity remains constant at 50\%). The plot shows that accuracy decreases 
with increasing net sparsity for BERT-Base. However, for BERT-Tiny, accuracy increases and decreases as we increase 
net sparsity.

\begin{figure}
    \centering
    \includegraphics[width=0.9\linewidth]{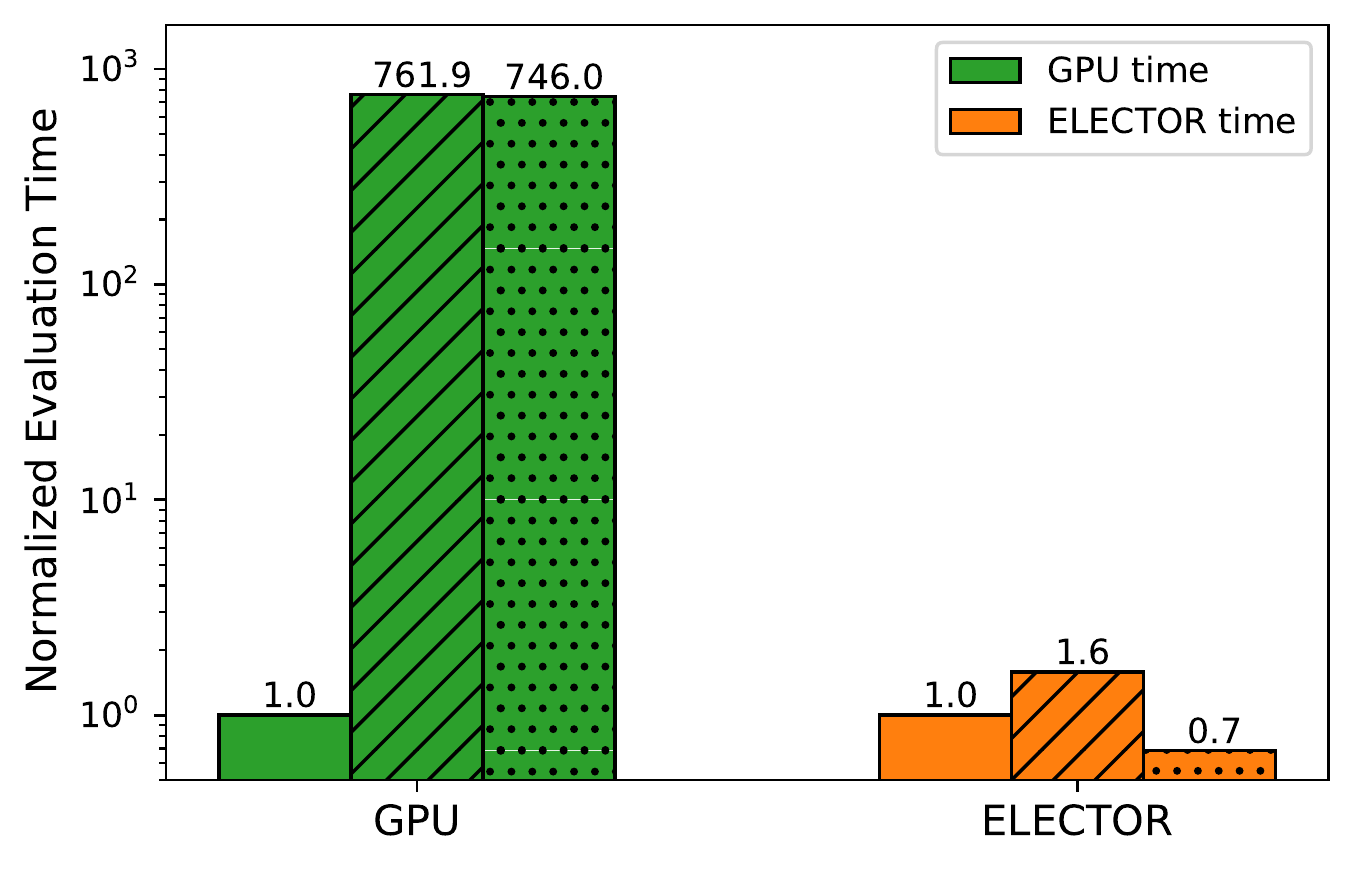}
    \caption{Evaluation time for traditional (hatched bars) and DynaProp (dotted bars) training normalized by the 
inference time (solid bars) on a GPU and an ELECTOR accelerator.}
    \label{fig:norm_time}
\end{figure}

Fig.~\ref{fig:norm_time} shows a plot of the normalized time for traditional and DynaProp training. Here, we evaluate 
the BERT-Tiny model on an Nvidia A100 GPU and an ELECTOR-supported accelerator (AccelTran-Edge~\cite{acceltran} with 
added training support). Training takes 761.9$\times$ longer than inference on a GPU. However, ELECTOR only requires 
1.6$\times$ more time. This is due to optimized scheduling, tiling of operation matrices, specialized hardware modules, 
and a dataflow curated for transformer workflows~\cite{acceltran}. Since an off-the-shelf GPU does not automatically 
skip ineffectual computations (in other words, it is not \emph{sparsity-aware}), DynaProp training hardly reduces 
evaluation time on the A100 GPU. However, due to the zero-free data format and specially designed hardware modules 
that skip ineffectual operations, ELECTOR reduces the training time by 2.3$\times$. Thus, high activation, weight, 
and gradient sparsities enabled by DynaProp, along with ASIC-based acceleration, allow ELECTOR to substantially 
reduce evaluation times relative to a baseline GPU.

\subsection{Design Space Exploration}

\begin{figure}
    \centering
    \includegraphics[width=0.95\linewidth]{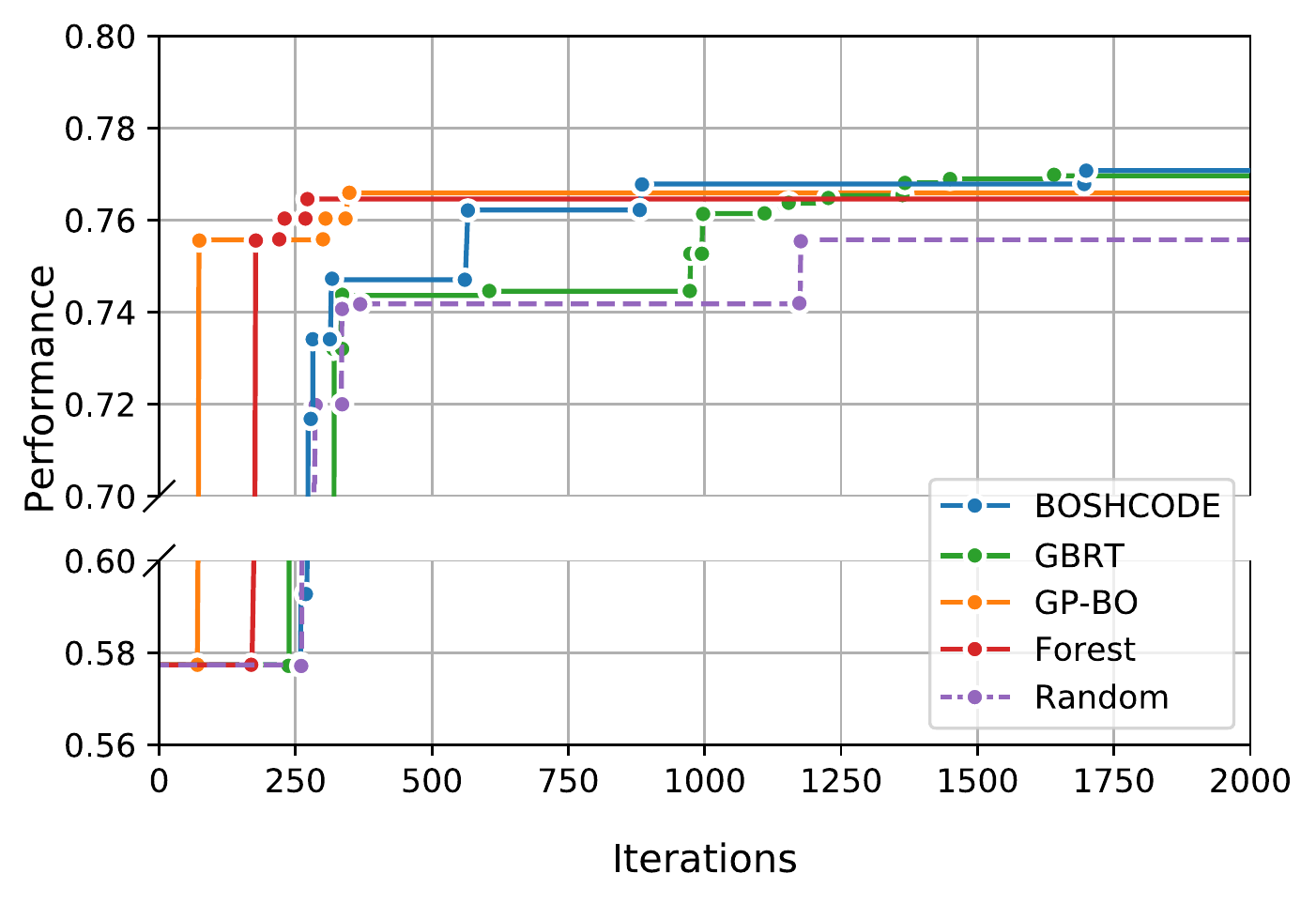}
    \caption{Co-design convergence plots for BOSHCODE and various baselines.}
    \label{fig:code_convergence}
\end{figure}

Fig.~\ref{fig:code_convergence} shows convergence plots while executing co-design using BOSHCODE and various baselines. 
These baselines include random search, gradient-boosted regression trees (GBRT), Gaussian-process-based Bayesian 
optimization (GP-BO) that approximates performance through Gaussian process regression and optimizes it through the 
L-BFGS method~\cite{lbfgs}, and random forest that fits various randomized decision trees over sub-samples of the dataset. 
As shown in Fig.~\ref{fig:code_convergence}, BOSHCODE achieves the highest performance. It yields the optimal 
transformer-accelerator pair, FB*-ELECTOR* (FB is an acronym for FlexiBERT 2.0). Here, performance refers to the net 
measure found using a convex combination of accuracy, latency, area, dynamic energy, and leakage energy 
(Section~\ref{sec:transcode}).

\begin{table}[]
\caption{Design choices of the converged TransCODE pair.}
\centering
\begin{tabular}{@{}l@{\hskip 0.2in}l@{\hskip 0.2in}|@{\hskip 0.2in}l@{}}
\toprule
\textbf{Hyperparameter}                   &        & \textbf{Value}            \\ \midrule
\multicolumn{3}{c}{\textbf{Transformer}} \\ \midrule
\multirow{8}{*}{Encoder Layer 1} & $h^1$  & 256                \\ [1mm]
                                 & \#SA-SDP & 3                \\ [1mm]
                                 & \#SA-WMA & 1                \\ [1mm]
                                 & \#LT-DFT & 1                \\ [1mm]
                                 & \#DSC-5  & 1                \\ [1mm]
                                 & \#DSC-9  & 1                \\ [1mm]
                                 & \#DSC-13 & 5                \\ [1mm]
                                 & FF     & 1024, 1024, 512    \\ \midrule
\multirow{6}{*}{Encoder Layer 2} & $h^2$  & 512                \\ [1mm]
                                 & \#LT-DFT & 4                \\ [1mm]
                                 & \#DSC-5  & 5                \\ [1mm]
                                 & \#DSC-9  & 3                \\ [1mm]
                                 & FF     & 256, 1024, 1024    \\ \midrule
\multicolumn{3}{c}{\textbf{Accelerator}} \\ \midrule
Batch tile size & & 4 \\
Spatial tile size & & 32 \\
Activation function & & GeLU \\
\#PEs & & 128 \\
\#MAC lanes per PE & & 32 \\
\#MACs per lane & & 16 \\
\#Softmax modules per PE & & 4 \\
Batch size & & 4 \\
Act./grad. buffer size (MB) & & 64 \\
Weight buffer size (MB) & & 128 \\
Mask buffer size (MB) & & 8 \\
Main memory configuration & & RRAM [8, 2, 4] \\ \bottomrule
\end{tabular}
\label{tbl:trancode_design_choices}
\end{table}

Table~\ref{tbl:trancode_design_choices} summarizes the design choices of the converged co-design pair, i.e., 
FB*-ELECTOR*. To optimize latency, FB* uses only two encoder layers. However, FB* uses 12 attention heads in each 
encoder layer to avoid performance loss. Thus, BOSHCODE searches for a shallow but wide model to improve throughput 
while not incurring a performance penalty. The converged architecture is also highly heterogeneous, with diverse 
attention types in each layer, leveraging the modeling capabilities of each operation type. ELECTOR* has many PEs to 
parallelize the computation of 12 attention heads in each FB* layer. It also leverages monolithic-3D RRAM, which has the 
highest bandwidth and lowest energy consumption. The net area of this accelerator is 359.3 mm$^2$.

\subsection{Performance Improvements}

\begin{figure*}
    \centering
    \includegraphics[width=0.87\linewidth]{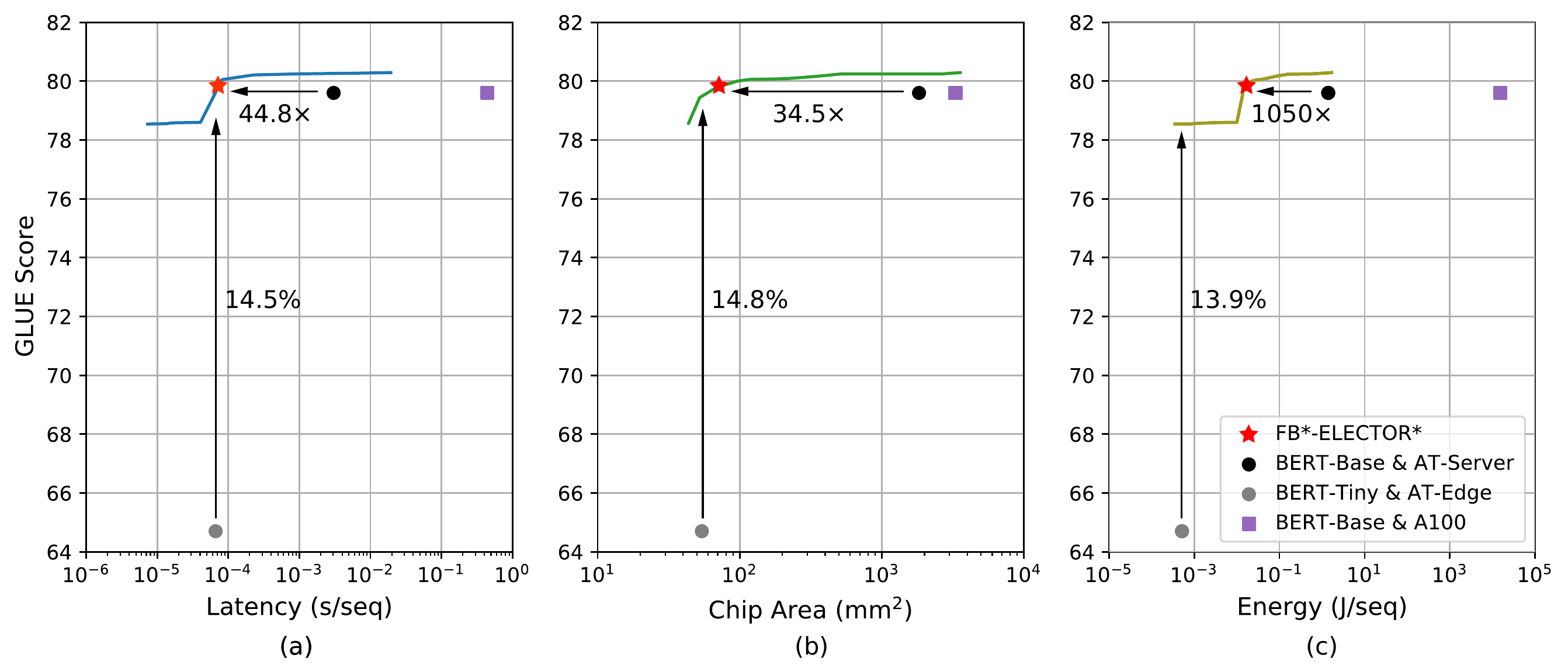}
    \caption{Pareto frontiers of GLUE scores for models in the FlexiBERT 2.0 design space with (a) latency, (b) chip area, and (c) energy consumption on difference accelerators in the ELECTOR design space. All performance values correspond to model training. AT stands for a modified version of AccelTran that supports training.}
    \label{fig:pareto}
\end{figure*}

\begin{table*}
\caption{Ablation analysis and baseline comparisons for our proposed TransCODE framework.}
\centering
\resizebox{\linewidth}{!}{
\begin{tabular}{@{}lcccccccc@{}}
\toprule
\textbf{Method}                  & \textbf{\begin{tabular}[c]{@{}c@{}}Hardware-\\ Aware\end{tabular}} & \textbf{\begin{tabular}[c]{@{}c@{}}Flex. \\ Layers\end{tabular}} & \textbf{\begin{tabular}[c]{@{}c@{}}Co-design\end{tabular}} & \textbf{\begin{tabular}[c]{@{}c@{}}ASIC-based \\ Accelerator\end{tabular}} & \textbf{Platform} & \textbf{\begin{tabular}[c]{@{}c@{}}GLUE Score \\ (\%)\end{tabular}} & \textbf{\begin{tabular}[c]{@{}c@{}}Latency \\ (ms/seq)\end{tabular}} & \textbf{\begin{tabular}[c]{@{}c@{}}Energy \\ (J/seq)\end{tabular}} \\ \midrule
Devlin et al.~\cite{bert} & \xmark & \xmark & \xmark & \xmark & A100 & 79.6 & 10.6 & 0.6 \\ \midrule
\multicolumn{9}{c}{\textbf{Baselines}} \\ \midrule
Wang et al.~\cite{hat_mit} & \cmark & \xmark & \xmark & \xmark & Raspberry Pi & 77.1 & 12,351.6 & 38.2 \\ [1mm] 
Yin et al.~\cite{autotinybert} & \xmark & \xmark & \xmark & \xmark & Raspberry Pi & 78.3 & 10,427.7 & 20.7  \\ [1mm]
Peng et al.~\cite{peng_dac_22} & \cmark & \xmark & \cmark & \xmark & FPGA & 77.0 & 15.8 & 0.4 \\ [1mm] 
Wang et al.~\cite{spatten} & \cmark & \xmark & \xmark & \cmark & SpAtten & 77.1 & 2.44 & 0.3 \\ [1mm]
Tuli et al.~\cite{acceltran} & \cmark & \xmark & \xmark & \cmark & AccelTran-Server & 79.6 & 0.26 & 0.06 \\ \midrule
\multicolumn{9}{c}{\textbf{Ablation Analysis}} \\ \midrule
\textbf{TransCODE (HW-NAS; Ours)} & \cmark & \cmark & \xmark & \cmark & ELECTOR & 78.4 & 0.23 & 0.08 \\ [1mm]
\textbf{TransCODE (without DynaProp; Ours)} & \cmark & \cmark & \cmark & \cmark & ELECTOR & 79.9 & 0.11 & 0.04 \\ [1mm]
\textbf{TransCODE (Ours)} & \cmark & \cmark & \cmark & \cmark & ELECTOR & \textbf{79.9} & \textbf{0.05} & \textbf{0.02} \\ \bottomrule
\end{tabular}}
\label{tbl:baseline_and_ablation}
\end{table*}

We now compare the converged transformer-accelerator pairs obtained by the proposed approach with baseline pairs. 
Fig.~\ref{fig:pareto} shows Pareto frontiers of GLUE scores with respect to hardware measures, i.e., latency, 
chip area, and energy consumption. We obtain GLUE scores from the surrogate model described in the EdgeTran 
framework~\cite{edgetran}. We also plot state-of-the-art transformer-accelerator pairs for comparison. Our pair 
on the Pareto frontier with the same accuracy as BERT-Base evaluated on AccelTran-Server incurs 44.8$\times$ 
lower latency. On the other hand, the pair on the Pareto frontier with the same latency as that of BERT-Tiny 
evaluated on AccelTran-Edge achieves a 14.5\% higher GLUE score. Similarly, the pair with the same accuracy as 
that of BERT-Base evaluated on AccelTran-Server but on the Pareto frontier in Fig.~\ref{fig:pareto}(b) requires 
34.5$\times$ lower chip area. The one with the same chip area as that evaluated on AccelTran-Edge finds a 
transformer model on the frontier that achieves a 14.8\% higher GLUE score. Finally, the pair with the same 
accuracy as that of BERT-Base incurs 1050$\times$ lower energy consumption than that of the model 
evaluated on AccelTran-Server. In contrast, the \emph{same-energy} pair with BERT-Tiny evaluated on 
AccelTran-Edge, but on the Pareto frontier, achieves a 13.9\% higher GLUE score.

Table~\ref{tbl:baseline_and_ablation} compares the proposed TransCODE approach against various baselines. 
These baselines include HAT~\cite{hat_mit} and AutoTinyBERT~\cite{autotinybert}, which implement HW-NAS on 
off-the-shelf edge-AI devices. We also add a co-design method implemented on a set of FPGAs~\cite{peng_dac_22} 
and another HW-NAS approach implemented on the SpAtten ASIC-based accelerator~\cite{spatten}. For fair 
comparisons, we also include a monolithic-3D-RRAM-based transformer accelerator, i.e., 
AccelTran-Server~\cite{acceltran}, that evaluates BERT-Base. Finally, the table presents an ablation study
in which we implement HW-NAS (by forcing the gradients to the accelerator to zero) with 
AccelTran-Server~\cite{acceltran} as the base accelerator. We also include performance values for 
FB*-ELECTOR* without DynaProp training implemented. Since the baselines do not support training, we report 
performance values for running inference with the proposed pairs. FB*-ELECTOR* outperforms the state-of-the-art 
pair, i.e., BERT-Base/AccelTran-Server, achieving 0.3\% higher accuracy, 5.2$\times$ lower latency, and 
3.0$\times$ lower energy consumption.

\section{\textcolor{black}{Discussions and Future Work}}
\label{sec:discussion}

\textcolor{black}{In this section, we discuss the implications of the proposed work along with future work directions.}

\subsection{\textcolor{black}{Multi-objective Optimization}}

\textcolor{black}{To perform co-design with the BOSHCODE framework, we model performance as a linear function of latency, energy 
consumption, chip area, and accuracy. This converts a multi-objective optimization problem into a single-objective optimization 
problem. We use this approach because BOSHCODE supports single-objective optimization only. The designer can decide the importance 
of each such objective when running the co-design pipeline. However, one could extend this approach to multi-objective optimization 
that increases/decreases a Pareto front's hypervolume~\cite{moo_rram, nsga2}. In this case, the designer would obtain a set of 
non-dominated solutions. We leave the application of multi-objective optimization methods to the FlexiBERT 2.0 and ELECTOR design 
spaces to future work.}

\subsection{\textcolor{black}{In-memory and Reconfigurable Processors}}

\textcolor{black}{The proposed framework optimizes for a specific accelerator deployed in practice for edge-based training or 
inference. However, any accelerator in the proposed ELECTOR design space can execute any transformer, although it would not be 
the best accelerator for that transformer (in terms of hardware performance). The hardware architectures are not reconfigurable 
at runtime (except the pruning ratios $\tau_\text{I}$ and $\tau_\text{T}$). The architectures in the Sanger~\cite{sanger} design 
space are reconfigurable. However, Sanger is limited to only pruning a given model in the software space. Meanwhile, 
TransCODE leverages the FlexiBERT 2.0 design space to search for dense and small models. It also supports dynamic
pruning of the model (using runtime-tunable pruning ratios) to trade off accuracy with hardware performance, while also searching 
for the best-performing set of accelerator design decisions. Nevertheless, adding reconfigurability to accelerators in the 
ELECTOR design space would benefit dynamic workloads. One could also implement co-design for a group of 
transformers instead of just one. We leave this to future work.}

\section{Conclusion}
\label{sec:conclusion}


In this work, we presented TransCODE, a co-design framework for flexible and heterogeneous transformer models 
evaluated on diverse accelerator architectures. We proposed a novel, low-overhead dynamic inference-and-training 
scheme, DynaProp, that increases the sparsity of activations and gradients at runtime with controllable accuracy 
loss. DynaProp attains 90\% sparsity in gradient matrices with negligible accuracy loss while improving training 
throughput by 2.3$\times$ relative to traditional training. We further proposed a design space of diverse 
ASIC-based transformer accelerators: ELECTOR. It supports accelerators targeted at various scenarios, budgets, 
and user-defined constraints that support flexible and heterogeneous transformer inference and training. The 
best transformer-accelerator pair achieves 0.3\% higher accuracy than the state-of-the-art pair while enabling 
5.2$\times$ lower latency and 3.0$\times$ lower energy consumption.

\section*{Acknowledgments}

We performed the simulations presented in this article on computational resources managed and supported by 
Princeton Research Computing at Princeton University.

\bibliographystyle{IEEEtran}
{\footnotesize
\bibliography{IEEEabrv, biblio}}

\begin{IEEEbiography}[{\includegraphics[width=1in,height=1.5in,clip,keepaspectratio]{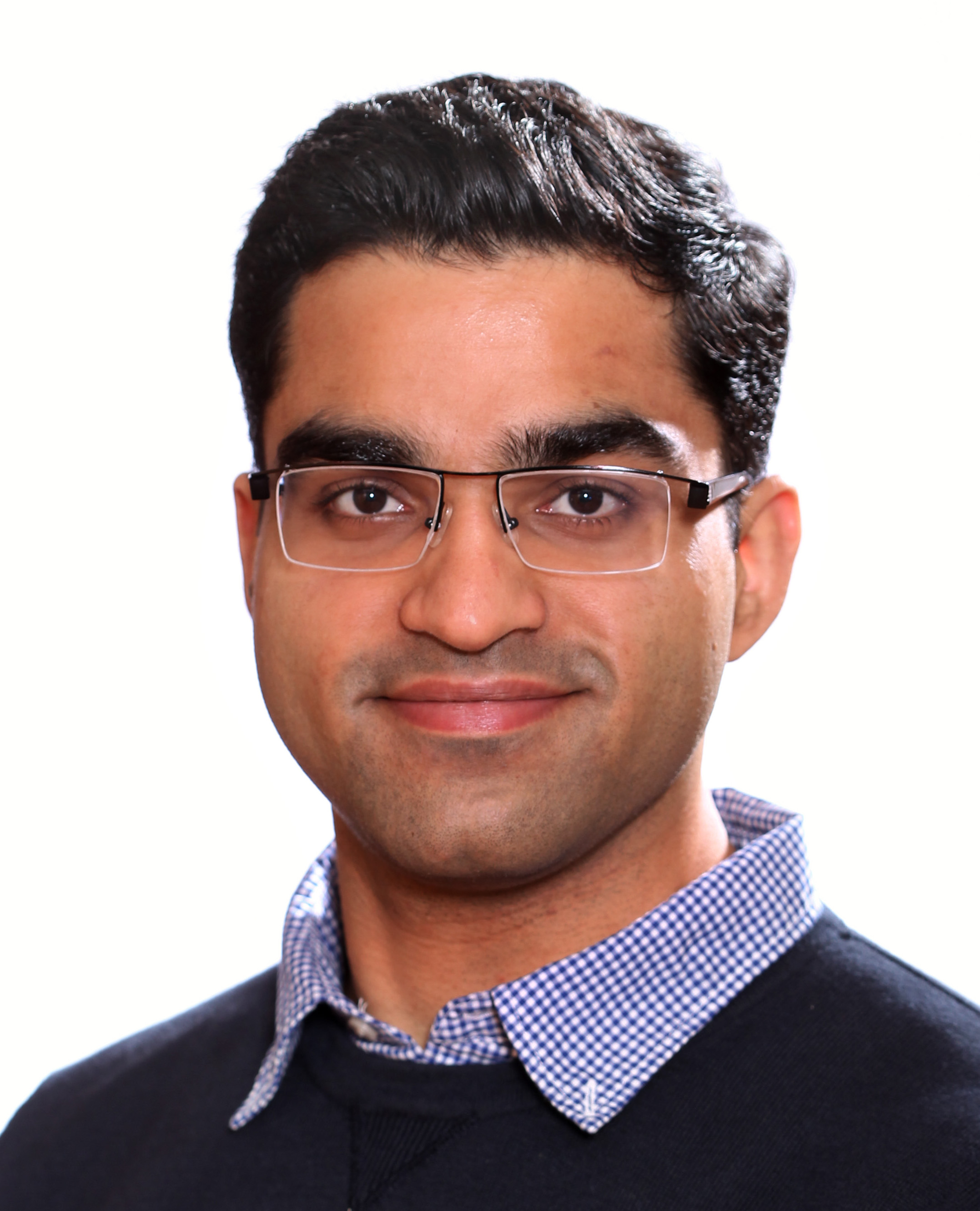}}]{Shikhar Tuli}
received the B. Tech. degree in electrical and electronics engineering from the Indian Institute of Technology (IIT) 
Delhi, India, with a department specialization in very large-scale integration (VLSI) and embedded systems. He is 
currently pursuing a Ph.D. degree at Princeton University in the department of electrical and computer engineering. 
His research interests include deep learning, edge artificial intelligence (AI), hardware-software co-design, 
brain-inspired computing, and smart healthcare.
\end{IEEEbiography}

\begin{IEEEbiography}[{\includegraphics[width=1in,height=1.5in,clip,keepaspectratio]{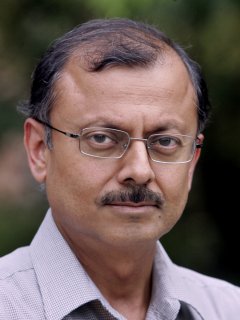}}]{Niraj K. Jha}
(Fellow, IEEE) received the B.Tech. degree in electronics and electrical communication engineering from 
IIT, Kharagpur, India, in 1981, and the Ph.D. degree in electrical engineering from the 
University of Illinois at Urbana–Champaign, Champaign, IL, USA, in 1985. 
He is a professor of electrical and computer engineering, Princeton University. 
He has co-authored five widely used books. He has published more than 470 papers (h-index: 83). 
He has received the Princeton Graduate Mentoring Award. His research has won 15 best paper awards, six award 
nominations, and 25 patents. He was given the Distinguished Alumnus Award by IIT, Kharagpur, in 2014. He has served 
as the Editor-in-Chief of TVLSI and an associate editor of several IEEE Transactions and other journals. He has 
given several keynote speeches in the areas of nanoelectronic design/test, smart healthcare, and cybersecurity.  
He is a fellow of ACM. His research interests include machine learning algorithms/architectures and smart healthcare. 
\end{IEEEbiography}

\end{document}